%% file: deep-limits.tex
\documentclass{article} 
\usepackage{nips14submit_e,times}
\usepackage{hyperref}
\usepackage{natbib}
\include{preamble}

\usepackage[margin=1.25in]{geometry}

\title{Avoiding pathologies in very deep networks}

\author{
David Duvenaud\\
University of Cambridge
\And
Oren Rippel \\
Harvard, M.I.T.
\And
Ryan P. Adams\\
Harvard University
\And
Zoubin Ghahramani \\
University of Cambridge
}

\nipsfinalcopy 

\begin{document}

\maketitle

\begin{abstract}
Choosing appropriate architectures and regularization strategies of deep networks is crucial to good predictive performance.
To shed light on this problem, we analyze the analogous problem of constructing useful priors on compositions of functions.
Specifically, we study the deep Gaussian process, a type of infinitely-wide, deep neural network.
We show that in standard architectures, the representational capacity of the network tends to capture fewer degrees of freedom as the number of layers increases, retaining only a single degree of freedom in the limit.
We propose an alternate network architecture which does not suffer from this pathology.
We also examine deep covariance functions, obtained by composing infinitely many feature transforms.
Lastly, we characterize the class of models obtained by performing dropout on Gaussian processes.
\end{abstract}

In this paper, we propose to study the problem of choosing neural net architectures by viewing deep neural networks as priors on functions.
By viewing neural networks this way, one can analyze their properties without reference to any particular dataset, loss function, or training method.

As a starting point, we will relate neural networks to Gaussian processes, and examine a class of infinitely-wide, deep neural networks called \emph{deep Gaussian processes}: compositions of functions drawn from \gp{} priors.
Deep \gp{}s are an attractive model class to study for several reasons.
First, \citet{damianou2012deep} showed that the probabilistic nature of deep \gp{}s guards against overfitting.
Second, \citet{hensman2014deep} showed that stochastic variational inference is possible in deep \gp{}s, allowing mini-batch training on large datasets.
Third, the availability of an approximation to the marginal likelihood allows one to automatically tune the model architecture without the need for cross-validation.
Finally, deep \gp{}s are attractive from a model-analysis point of view, because they abstract away some of the details of finite neural networks. 

Our analysis will show that in standard architectures, the representational capacity of standard deep networks tends to decrease as the number of layers increases, retaining only a single degree of freedom in the limit.
We propose an alternate network architecture that connects the input to each layer that does not suffer from this pathology.
We also examine \emph{deep kernels}, obtained by composing arbitrarily many fixed feature transforms.
Finally, we characterise the prior obtained by performing dropout regularization on \gp{}s, showing equivalences to existing models.


\section{Relating deep neural networks to deep \sgp{}s}
\label{sec:relating}

This section gives a precise definition of deep \gp{}s, reviews the precise relationship between neural networks and Gaussian processes, and gives two equivalent ways of constructing neural networks which give rise to deep \gp{}s.

\subsection{Definition of deep \sgp{}s}

We define a deep \gp{} as a distribution on functions constructed by composing functions drawn from \gp{} priors.
An example of a deep \gp{} is a composition of vector-valued functions, with each function drawn independently from \gp{} priors:
\begin{align}
\vf^{(1:L)}(\vx) = \vf^{(L)}(\vf^{(L-1)}(\dots \vf^{(2)}(\vf^{(1)}(\vx)) \dots)) \\
\textnormal{with each} \quad f_d^{(\layerindex)}  \simind \gp{} \left( 0, k^{(\layerindex)}_d(\vx, \vx') \right) \nonumber
\label{eq:deep-gp}
\end{align}
%

Multilayer neural networks also implement compositions of vector-valued functions, one per layer.
Therefore, understanding general properties of function compositions might helps us to gain insight into deep neural networks.

\subsection{Single-hidden-layer models}

First, we relate neural networks to standard ``shallow'' Gaussian processes, using the standard neural network architecture known as the multi-layer perceptron (\MLP{})~\citep{rosenblatt1962principles}.
In the typical definition of an \MLP{} with one hidden layer, the hidden unit activations are defined as:
\begin{align}
\vh(\vx) = \sigma \left( \vb + \munitparams \vx \right)
\end{align}
where $\vh$ are the hidden unit activations, $\vb$ is a bias vector, $\munitparams$ is a weight matrix and $\sigma$ is a one-dimensional nonlinear function, usually a sigmoid, applied element-wise. The output vector $\vf(\vx)$ is simply a weighted sum of these hidden unit activations:
\begin{align}
\vf(\vx) = \mnetweights \sigma \left( \vb + \munitparams \vx \right)  = \mnetweights \vh(\vx)
\label{eq:one-layer-nn}
\end{align}
where $\mnetweights$ is another weight matrix.

\citet[chapter 2]{neal1995bayesian} showed that some neural networks with infinitely many hidden units, one hidden layer, and unknown weights correspond to Gaussian processes.
More precisely, for any model of the form
\begin{align}
f(\vx) = \frac{1}{K}{\mathbf \vnetweights}\tra \hPhi(\vx) = \frac{1}{K} \sum_{i=1}^K \netweights_i \hphi_i(\vx),
\label{eq:one-layer-gp}
\end{align}
with fixed%
 features $\left[ \hphi_1(\vx), \dots, \hphi_K(\vx) \right]\tra = \hPhi(\vx)$ and i.i.d. $\netweights$'s with zero mean and finite variance $\sigma^2$, the central limit theorem implies that as the number of features $K$ grows, any two function values $f(\vx)$ and $f(\vx')$ have a joint distribution approaching a Gaussian:
\begin{align}
\lim_{K \to \infty} p\left( \colvec{f(\vx)}{f(\vx')} \right) = \Nt{\colvec{0}{0}}{
\frac{\sigma^2}{K} \left[ \begin{array}{cc}
\sum_{i=1}^K \hphi_i(\vx)\hphi_i(\vx) &
\sum_{i=1}^K \hphi_i(\vx)\hphi_i(\vx') \\
\sum_{i=1}^K \hphi_i(\vx')\hphi_i(\vx) &
\sum_{i=1}^K \hphi_i(\vx')\hphi_i(\vx')
\end{array} \right] }
\end{align}
A joint Gaussian distribution between any set of function values is the definition of a Gaussian process.

The result is surprisingly general:
it puts no constraints on the features (other than having uniformly bounded activation), nor does it require that the feature weights $\vnetweights$ be Gaussian distributed.  
An \MLP{} with a finite number of nodes also gives rise to a \gp{}, but only if the distribution on $\vnetweights$ is Gaussian.

One can also work backwards to derive a one-layer \MLP{} from any \gp{}:
Mercer's theorem, discussed in \cref{sec:mercer}, implies that any positive-definite kernel function corresponds to an inner product of features: $k(\vx, \vx') = \hPhi(\vx) \tra \hPhi(\vx')$.

Thus, in the one-hidden-layer case, the correspondence between \MLP{}s and \gp{}s is straightforward:
the implicit features $\hPhi(\vx)$ of the kernel correspond to hidden units of an \MLP{}.

\newcommand{\numdims}[0]{3}
\newcommand{\numhidden}[0]{3}
\newcommand{\upnodedist}[0]{1cm}
\newcommand{\bardist}[0]{\hspace{-0.2cm}}

\def\layersep{2.3cm}
\def\nodesep{1.3cm}
\def\nodesize{1cm}

\newcommand{\neuronfunc}[2]{
\FPeval{\result}{clip(#1+#2)}
\includegraphics[width=1cm, clip, trim=0mm 0mm 0mm 0mm]{figures/deep-limits/two-d-draws/sqexp-draw-\result}
}

\tikzstyle{input neuron}=[neuron, fill=green!15]
\tikzstyle{output neuron}=[neuron, fill=red!15]
\tikzstyle{hidden neuron}=[neuron, fill=blue!15]

\newcommand{\indfeat}{h}

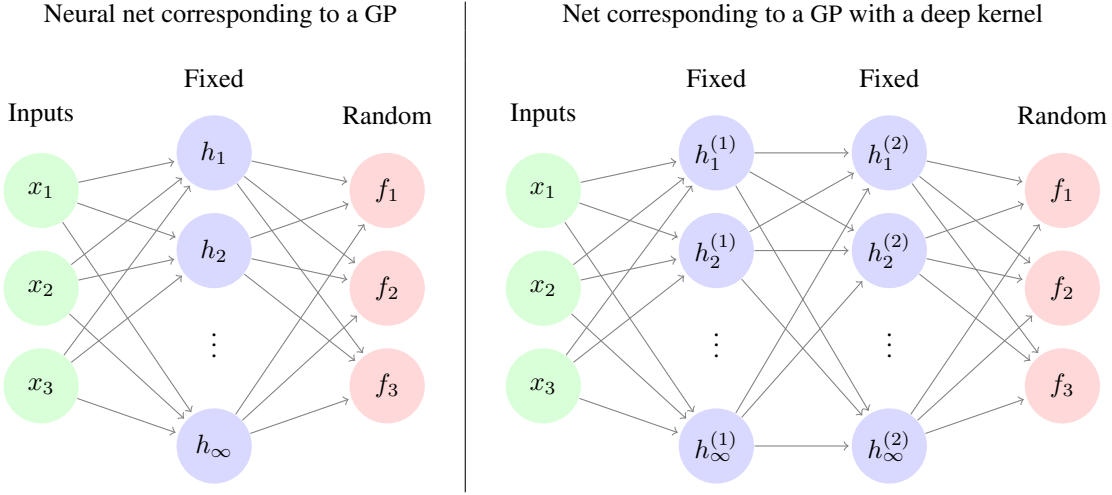
\begin{figure}[t]
\begin{tabular}{c|c}
Neural net corresponding to a \gp{} & Net corresponding to a \gp{} with a deep kernel \\
\\
\null\hspace{-0.25cm}
\begin{tikzpicture}[shorten >=1pt,->,draw=black!50, node distance=\layersep]
    \tikzstyle{every pin edge}=[<-,shorten <=1pt]
    \tikzstyle{neuron}=[circle,fill=black!25,minimum size=17pt,inner sep=0pt]
    \tikzstyle{annot} = [text width=4em, text centered]

    \foreach \name / \y in {1,...,\numdims}
        \node[input neuron, minimum size=\nodesize
        ] (I-\name) at (0,-\nodesep*\y) {$x_\y$};

    \foreach \name / \y in {1,2}
        \path[yshift=0.5cm]
            node[hidden neuron, minimum size=\nodesize] (H-\name) at (\layersep,-\nodesep*\y) {$\indfeat_{\y}$};
    
	\foreach \name / \y in {3}
	    \path[yshift=0.5cm]
    	    node[hidden neuron, minimum size=\nodesize] (H-\name) at (\layersep,-\nodesep*4) {$\indfeat_\infty$};

    \foreach \name / \y in {1,...,\numdims}
    	\node[output neuron, minimum size=\nodesize] (O-\name) at (2*\layersep,-\nodesep*\y) {$f_{\y}$};

    \foreach \source in {1,...,\numdims}
        \foreach \dest in {1,...,\numhidden}
            \path (I-\source) edge (H-\dest);

    \foreach \source in {1,...,\numhidden}
        \foreach \dest in {1,...,\numdims}
    	    \path (H-\source) edge (O-\dest);

    \node[annot,below of=H-2, node distance=1.15cm] {$\vdots$};
    \node[annot,above of=I-1, node distance=\upnodedist] {Inputs};
    \node[annot,above of=H-1, node distance=\upnodedist] {Fixed};
    \node[annot,above of=O-1, node distance=\upnodedist] {Random};
\end{tikzpicture}
&
\begin{tikzpicture}[shorten >=1pt,->,draw=black!50, node distance=\layersep]
    \tikzstyle{every pin edge}=[<-,shorten <=1pt]
    \tikzstyle{neuron}=[circle,fill=black!25,minimum size=17pt,inner sep=0pt]
    \tikzstyle{annot} = [text width=4em, text centered]

    \foreach \name / \y in {1,...,\numdims}
        \node[input neuron, minimum size=\nodesize
        ] (I-\name) at (0,-\nodesep*\y) {$x_\y$};

    \foreach \name / \y in {1,2}
        \path[yshift=0.5cm]
            node[hidden neuron, minimum size=\nodesize] (H-\name) at (\layersep,-\nodesep*\y) {$\indfeat^{(1)}_{\y}$};
	\foreach \name / \y in {3}
	    \path[yshift=0.5cm]
    	    node[hidden neuron, minimum size=\nodesize] (H-\name) at (\layersep,-\nodesep*4) {$\indfeat^{(1)}_\infty$};

    \foreach \name / \y in {1,2}
        \path[yshift=0.5cm]
            node[hidden neuron, minimum size=\nodesize] (H2-\name) at (2*\layersep,-\nodesep*\y) {$\indfeat^{(2)}_{\y}$};
	\foreach \name / \y in {3}
	    \path[yshift=0.5cm]
    	    node[hidden neuron, minimum size=\nodesize] (H2-\name) at (2*\layersep,-\nodesep*4) {$\indfeat^{(2)}_\infty$};

    \foreach \name / \y in {1,...,\numdims}
    	\node[output neuron, minimum size=\nodesize
    	] (O-\name) at (3*\layersep,-\nodesep*\y) {$f_{\y}$};

    \foreach \source in {1,...,\numdims}
        \foreach \dest in {1,...,\numhidden}
            \path (I-\source) edge (H-\dest);
            
    \foreach \source in {1,...,\numhidden}
        \foreach \dest in {1,...,\numhidden}
            \path (H-\source) edge (H2-\dest);            

    \foreach \source in {1,...,\numhidden}
        \foreach \dest in {1,...,\numdims}
    	    \path (H2-\source) edge (O-\dest);

    \node[annot,above of=I-1, node distance=\upnodedist] {Inputs};
    \node[annot,below of=H-2, node distance=1.15cm] {$\vdots$};    
    \node[annot,above of=H-1, node distance=\upnodedist] {Fixed};
    \node[annot,below of=H2-2, node distance=1.15cm] {$\vdots$};
    \node[annot,above of=H2-1, node distance=\upnodedist] {Fixed};
    \node[annot,above of=O-1, node distance=\upnodedist] {Random};
\end{tikzpicture}
\end{tabular}
\caption[Neural network architectures giving rise to \sgp{}s]
{
\emph{Left:} \gp{}s can be derived as a one-hidden-layer \MLP{} with infinitely many fixed hidden units having unknown weights.
\emph{Right:} Multiple layers of fixed hidden units gives rise to a \gp{} with a deep kernel, but not a deep \gp{}.
}
\label{fig:gp-architectures}
\end{figure}

\subsection{Multiple hidden layers}
Next, we examine infinitely-wide \MLP{}s having multiple hidden layers.
There are several ways to construct such networks, giving rise to different priors on functions.

In an \MLP{} with multiple hidden layers, activation of the $\layerindex$th layer units are given by
%
\begin{align}
\vh^{(\layerindex)}(\vx) = \sigma \left( \vb^{(\layerindex)} + \munitparams^{(\layerindex)} \vh^{(\layerindex-1)}(\vx) \right) \;.
\label{eq:nextlayer}
\end{align}
This architecture is shown on the right of \cref{fig:gp-architectures}.
%
For example, if we extend the model given by \cref{eq:one-layer-nn} to have two layers of feature mappings, the resulting model becomes
\begin{align}
f(\vx) = \frac{1}{K}{\mathbf \vnetweights}\tra \hPhi^{(2)}\left( \hPhi^{(1)}(\vx) \right) \;.
\label{eq:mutli-layer-nn}
\end{align}

If the features $\vh^{(1)}(\vx)$ and $\vh^{(2)}(\vx)$ are fixed with only the last-layer weights ${\vnetweights}$ unknown, this model corresponds to a shallow \gp{} with a \emph{deep kernel}, given by
\begin{align}
k(\vx, \vx') = \left[ \hPhi^{(2)} ( \hPhi^{(1)}(\vx) ) \right] \tra \hPhi^{(2)} (\hPhi^{(1)}(\vx') ) \, .
\end{align}

Deep kernels, explored in \cref{sec:deep_kernels}, imply a fixed representation as opposed to a prior over representations.
Thus, unless we richly parameterize these kernels, their capacity to learn an appropriate representation will be limited in comparison to more flexible models such as deep neural networks or deep \gp{}s. 

\subsection{Two network architectures equivalent to deep \sgp{}s}

\def\halfshift{0.0cm}

\begin{figure}[t!]
\centering
\begin{tabular}{c}
A neural net with fixed activation functions corresponding to a 3-layer deep \gp{}\\ 
\\
\begin{tikzpicture}[shorten >=1pt,->,draw=black!50, node distance=\layersep]
    \tikzstyle{every pin edge}=[<-,shorten <=1pt]
    \tikzstyle{neuron}=[circle,fill=black!25,minimum size=17pt,inner sep=0pt]
    \tikzstyle{annot} = [text width=4em, text centered]

    \foreach \name / \y in {1,...,\numdims}
        \node[input neuron, minimum size=\nodesize
        ] (I-\name) at (0,-\nodesep*\y) {$x_\y$};

    \foreach \name / \y in {1,2}
        \path[yshift=0.5cm]
            node[hidden neuron, minimum size=\nodesize] (H-\name) at (\layersep,-\nodesep*\y) {$\indfeat^{(1)}_{\y}$};
   	\foreach \name / \y in {3}
	    \path[yshift=0.5cm]
    	    node[hidden neuron, minimum size=\nodesize] (H-\name) at (\layersep,-\nodesep*4) {$\indfeat^{(1)}_\infty$};

    \foreach \name / \y in {1,2}
        \path[yshift=0.5cm]
            node[hidden neuron, minimum size=\nodesize] (H2-\name) at (3*\layersep,-\nodesep*\y) {$\indfeat^{(2)}_{\y}$};
   	\foreach \name / \y in {3}
	    \path[yshift=0.5cm]
    	    node[hidden neuron, minimum size=\nodesize] (H2-\name) at (3*\layersep,-\nodesep*4) {$\indfeat^{(2)}_\infty$};
    	    
    \foreach \name / \y in {1,2}
        \path[yshift=0.5cm]
            node[hidden neuron, minimum size=\nodesize] (H3-\name) at (5*\layersep,-\nodesep*\y) {$\indfeat^{(3)}_{\y}$};
   	\foreach \name / \y in {3}
	    \path[yshift=0.5cm]
    	    node[hidden neuron, minimum size=\nodesize] (H3-\name) at (5*\layersep,-\nodesep*4) {$\indfeat^{(3)}_\infty$};    	    

    \foreach \name / \y in {1,...,\numdims}
    	\node[output neuron, minimum size=\nodesize
    	] (O1-\name) at (2*\layersep,-\nodesep*\y) {$f^{(1)}_{\y}$};

    \foreach \name / \y in {1,...,\numdims}
    	\node[output neuron, minimum size=\nodesize
    	] (O2-\name) at (4*\layersep,-\nodesep*\y) {$f^{(2)}_{\y}$};
    	
    \foreach \name / \y in {1,...,\numdims}
    	\node[output neuron, minimum size=\nodesize
    	] (O3-\name) at (6*\layersep,-\nodesep*\y) {$f^{(3)}_{\y}$};    	

    \foreach \source in {1,...,\numdims}
        \foreach \dest in {1,...,\numhidden}
            \path (I-\source) edge (H-\dest);
            
    \foreach \source in {1,...,\numhidden}
        \foreach \dest in {1,...,\numdims}
            \path (H-\source) edge (O1-\dest);         
            
    \foreach \source in {1,...,\numdims}
        \foreach \dest in {1,...,\numhidden}
            \path (O1-\source) edge (H2-\dest);        
            
    \foreach \source in {1,...,\numdims}
        \foreach \dest in {1,...,\numhidden}
            \path (O2-\source) edge (H3-\dest);                      
            
    \foreach \source in {1,...,\numhidden}
        \foreach \dest in {1,...,\numdims}
    	    \path (H3-\source) edge (O3-\dest);            

    \foreach \source in {1,...,\numhidden}
        \foreach \dest in {1,...,\numdims}
    	    \path (H2-\source) edge (O2-\dest);

    \node[annot,above of=I-1, node distance=\upnodedist] {Inputs};
    \node[annot,below of=I-3, node distance=\upnodedist] {$\vx$};
    \node[annot,above of=H-1, node distance=\upnodedist] {Fixed};
    \node[annot,below of=O1-3, node distance=\upnodedist] {$\vf^{(1)}(\vx)$};
    \node[annot,above of=O1-1, node distance=\upnodedist] {Random};
    \node[annot,above of=H2-1, node distance=\upnodedist] {Fixed};
    \node[annot,below of=O2-3, node distance=\upnodedist] {$\vf^{(1:2)}(\vx)$};
    \node[annot,above of=O2-1, node distance=\upnodedist] {Random};
    \node[annot,above of=H3-1, node distance=\upnodedist] {Fixed};
    \node[annot,above of=O3-1, node distance=\upnodedist] {Random};
    \node[annot,below of=O3-3, node distance=\upnodedist] {$\vy$};
    \node[annot,below of=H-2, node distance=1.15cm] {$\vdots$};    
    \node[annot,below of=H2-2, node distance=1.15cm] {$\vdots$};    
    \node[annot,below of=H3-2, node distance=1.15cm] {$\vdots$}; 
\end{tikzpicture}
\\[0.5em]
\hline
\\[-0.5em]
A net with nonparametric activation functions corresponding to a 3-layer deep \gp{}\\
\\
\begin{tikzpicture}[shorten >=1pt,->,draw=black!50, node distance=\layersep]
    \tikzstyle{every pin edge}=[<-,shorten <=1pt]
    \tikzstyle{neuron}=[circle, draw = black, inner sep=0pt, line width = 1pt]
    \tikzstyle{input neuron}=[circle, minimum size=\nodesize, fill=green!15]
    \tikzstyle{output neuron}=[neuron, minimum size=\nodesize, text width = 1cm];
    \tikzstyle{hidden neuron}=[neuron, minimum size=\nodesize, text width = 1cm];
    \tikzstyle{annot} = [text width=4em, text centered]

    \foreach \name / \y in {1,...,\numdims}
        \node[input neuron] (I-\name) at (0,-\nodesep*\y) {$x_\y$};

    \foreach \name / \y in {1,...,\numhidden}
        \path[yshift=\halfshift] node[hidden neuron] (H-\name) at (\layersep,-\nodesep*\y) { \neuronfunc{\y}{0}};

    \foreach \name / \y in {1,...,\numhidden}
        \path[yshift=\halfshift] node[hidden neuron] (H2-\name) at (2*\layersep,-\nodesep*\y) {\neuronfunc{\y}{4}};

    \foreach \name / \y in {1,...,\numdims}
    	\node[output neuron] (O-\name) at (3*\layersep,-\nodesep*\y) {\neuronfunc{\y}{8}};

    \foreach \source in {1,...,\numdims}
        \foreach \dest in {1,...,\numhidden}
            \path (I-\source) edge (H-\dest);
            
    \foreach \source in {1,...,\numhidden}
        \foreach \dest in {1,...,\numhidden}
            \path (H-\source) edge (H2-\dest);            

    \foreach \source in {1,...,\numhidden}
        \foreach \dest in {1,...,\numdims}
    	    \path (H2-\source) edge (O-\dest);

    \node[annot,above of=I-1, node distance=\upnodedist] {Inputs};
    \node[annot,below of=I-\numdims, node distance=\upnodedist] {$\vx$};    
    \node[annot,above of=H-1, node distance=\upnodedist, text width = 2cm] {\gp{}};
    \node[annot,above of=H2-1, node distance=\upnodedist, text width = 2cm] {\gp{}};
    \node[annot,below of=H-\numhidden, node distance=\upnodedist, text width = 2cm] {$\vf^{(1)}(\vx)$};
    \node[annot,below of=H2-\numhidden, node distance=\upnodedist, text width = 2cm] {$\vf^{(1:2)}(\vx)$};
    \node[annot,above of=O-1, node distance=\upnodedist] {\gp{}};
    \node[annot,below of=O-\numdims, node distance=\upnodedist, text width = 1cm] {$\vy$};
\end{tikzpicture}
\end{tabular}
\caption[Neural network architectures giving rise to deep \sgp{}s]
{
Two equivalent views of deep \gp{}s as neural networks.
\emph{Top:} A neural network whose every other layer is a weighted sum of an infinite number of fixed hidden units, whose weights are initially unknown.
\emph{Bottom:} A neural network with a finite number of hidden units, each with a different unknown non-parametric activation function.
The activation functions are visualized by draws from 2-dimensional \gp{}s, although their input dimension will actually be the same as the output dimension of the previous layer.
}
\label{fig:deep-gp-architectures}
\end{figure}
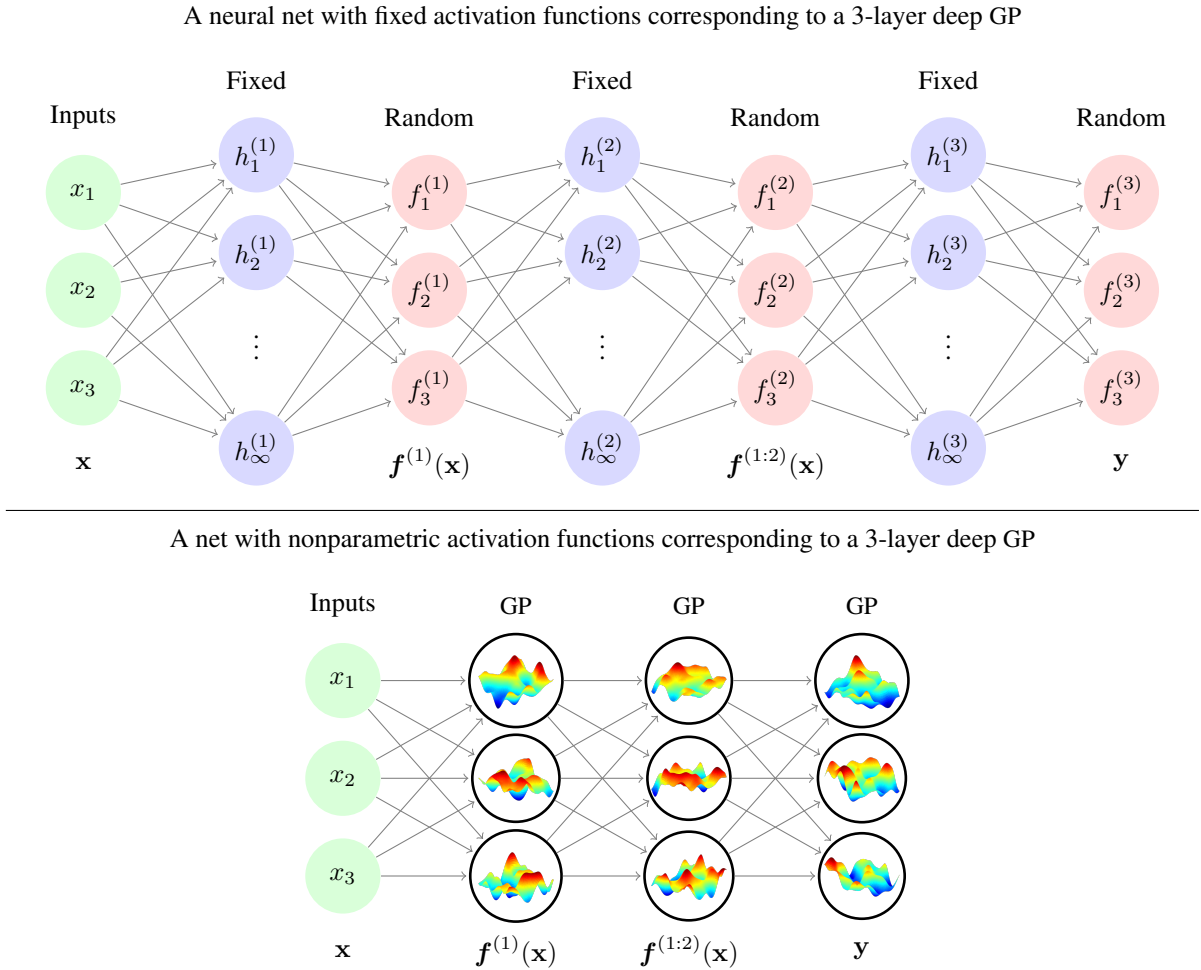

There are two equivalent neural network architectures that correspond to deep \gp{}s: one having fixed nonlinearities, and another having \gp{}-distributed nonlinearities.

To construct a neural network corresponding to a deep \gp{} using only fixed nonlinearities, 
one can start with the infinitely-wide deep \gp{} shown in \cref{fig:gp-architectures}(right), and introduce a finite set of nodes in between each infinitely-wide set of fixed basis functions.
This architecture is shown in the top of \cref{fig:deep-gp-architectures}.
The $D^{(\layerindex)}$ nodes $\vf^{(\layerindex)}(\vx)$ in between each fixed layer are weighted sums (with random weights) of the fixed hidden units of the layer below, and the next layer's hidden units depend only on these $D^{(\layerindex)}$ nodes.

This alternating-layer architecture has an interpretation as a series of linear information bottlenecks.
To see this, substitute \cref{eq:one-layer-nn} into \cref{eq:nextlayer} to get
\begin{align}
\hPhi^{(\layerindex)}(\vx) = \sigma \left( \vb^{(\layerindex)} + \left[ \munitparams^{(\layerindex)} \mnetweights^{(\layerindex-1)} \right] \hPhi^{(\layerindex-1)}(\vx) \right)
\end{align}
where $\mnetweights^{(\ell-1)}$ is the weight matrix connecting $\hPhi^{(\ell-1)}$ to $\vf^{(\ell - 1)}$.
Thus, ignoring the intermediate outputs $\vf^{(\layerindex)}(\vx)$, a deep \gp{} is an infinitely-wide, deep \MLP{} with each pair of layers connected by random, rank-$D_\layerindex$ matrices given by $\munitparams^{(\layerindex)} \mnetweights^{(\layerindex-1)}$.

The second, more direct way to construct a network architecture corresponding to a deep \gp{} is to integrate out all $\mnetweights^{(\layerindex)}$, and view deep \gp{}s as a neural network with a finite number of nonparametric, \gp{}-distributed basis functions at each layer, in which $\vf^{(1:\layerindex)}(\vx)$ represent the output of the hidden nodes at the $\layerindex^{th}$ layer.
This second view lets us compare deep \gp{} models to standard neural net architectures more directly.
\Cref{fig:gp-architectures}(bottom) shows an example of this architecture.

\section{Characterizing deep Gaussian process priors}
\label{sec:characterizing-deep-gps}

This section develops several theoretical results characterizing the behavior of deep \gp{}s as a function of their depth.
Specifically, we show that the size of the derivative of a one-dimensional deep \gp{} becomes log-normal distributed as the network becomes deeper.
We also show that the Jacobian of a multivariate deep \gp{} is a product of independent Gaussian matrices having independent entries.
These results will allow us to identify a pathology that emerges in very deep networks in \cref{sec:formalizing-pathology}.

\subsection{One-dimensional asymptotics}
\label{sec:1d}


In this section, we derive the limiting distribution of the derivative of an arbitrarily deep, one-dimensional \gp{} having a squared-exp kernel:  
\newcommand{\onedsamplepic}[1]{
\includegraphics[trim=2mm 5mm 7mm 6.4mm, clip, width=0.235\columnwidth]{\deeplimitsfiguresdir/1d_samples/latent_seed_0_1d_large/layer-#1}}%
\newcommand{\onedsamplepiccon}[1]{
\includegraphics[trim=2mm 5mm 7mm 6.4mm, clip, width=0.235\columnwidth]{\deeplimitsfiguresdir/1d_samples/latent_seed_0_1d_large_connected/layer-#1}}%
\begin{figure}
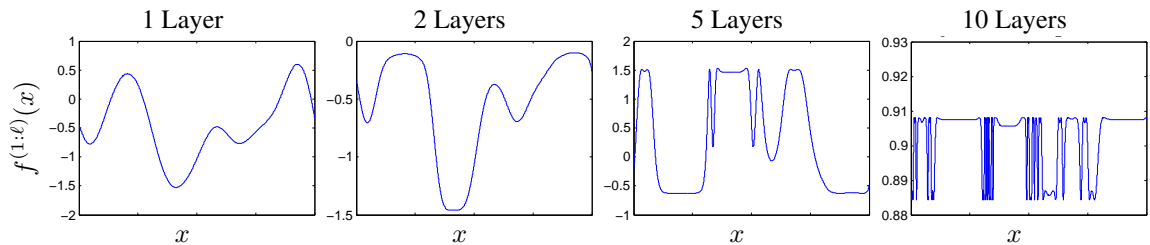

\centering
\setlength{\tabcolsep}{1.5pt}
\begin{tabular}{ccccc}
& 1 Layer & 2 Layers & 5 Layers & 10 Layers \\
\raisebox{0.6cm}{\rotatebox{90}{$f^{(1:\ell)}(x)$}} &
\onedsamplepic{1} &
\onedsamplepic{2} &
\onedsamplepic{5} &
\onedsamplepic{10} \\[-3pt]
 & $x$ & $x$ & $x$ & $x$
\end{tabular}
\caption[A one-dimensional draw from a deep \sgp{} prior]
{A function drawn from a one-dimensional deep \gp{} prior, shown at different depths.
The $x$-axis is the same for all plots.
After a few layers, the functions begin to be either nearly flat, or quickly-varying, everywhere.
This is a consequence of the distribution on derivatives becoming heavy-tailed.
As well, the function values at each layer tend to cluster around the same few values as the depth increases.
This happens because once the function values in different regions are mapped to the same value in an intermediate layer, there is no way for them to be mapped to different values in later layers.}
\label{fig:deep_draw_1d}
\end{figure}
\begin{align}
\kSE(x, x') = \sigma^2 \exp \left( \frac{-(x - x')^2}{2\lengthscale^2} \right) \;.
\label{eq:se_kernel}
\end{align}
The parameter $\sigma^2$ controls the variance of functions drawn from the prior, and the lengthscale parameter $\lengthscale$ controls the smoothness.  
The derivative of a \gp{} with a squared-exp kernel is point-wise distributed as $\Nt{0}{\nicefrac{\sigma^2}{\lengthscale^2}}$.  
Intuitively, a draw from a \gp{} is likely to have large derivatives if the kernel has high variance and small lengthscales.
 
By the chain rule, the derivative of a one-dimensional deep \gp{} is simply a product of the derivatives of each layer, which are drawn independently by construction.
The distribution of the absolute value of this derivative is a product of half-normals, each with mean $\sqrt{\nicefrac{2 \sigma^2}{\pi \lengthscale^2}}$.
%
%
If one chooses kernel parameters such that ${\nicefrac{\sigma^2}{\lengthscale^2} = \nicefrac{\pi}{2}}$, then the expected magnitude of the derivative remains constant regardless of the depth.


The distribution of the log of the magnitude of the derivatives has finite moments:
\begin{align}
m_{\log} = \expectargs{}{\log \left| \frac{\partial f(x)}{\partial x} \right|} & = 2 \log \left( \frac{\sigma}{\lengthscale} \right) - \log2 - \gamma \nonumber \\
v_{\log} = \varianceargs{}{\log \left| \frac{\partial f(x)}{\partial x} \right|} & = \frac{\pi^2}{4} + \frac{\log^2 2}{2}  - \gamma^2 - \gamma \log4 + 2 \log\left(\frac{\sigma}{\lengthscale}\right) \left[ \gamma + \log2 - \log \left(\frac{\sigma}{\lengthscale}\right) \right]
\end{align}
where $\gamma \approxeq 0.5772$ is Euler's constant.  Since the second moment is finite, by the central limit theorem, the limiting distribution of the size of the gradient approaches a log-normal as L grows:
\begin{align}
\log \left| \frac{\partial f^{(1:L)}(x)}{\partial x} \right| 
 = \log \prod_{\layerindex=1}^L \left| \frac{\partial f^{(\layerindex)}(x)}{\partial x} \right| 
 = \sum_{\layerindex=1}^L \log \left| \frac{\partial f^{(\layerindex)}(x)}{\partial x} \right| 
\distas{L \rightarrow \infty} \Nt{ L m_{\log} }{L^2 v_{\log}}
\end{align}
Even if the expected magnitude of the derivative remains constant, the variance of the log-normal distribution grows without bound as the depth increases.

Because the log-normal distribution is heavy-tailed and its domain is bounded below by zero, the derivative will become very small almost everywhere, with rare but very large jumps.  
\Cref{fig:deep_draw_1d} shows this behavior in a draw from a 1D deep \gp{} prior.
This figure also shows that once the derivative in one region of the input space becomes very large or very small, it is likely to remain that way in subsequent layers.
%

\subsection{Distribution of the Jacobian}
\label{sec:theorem}
Next, we characterize the distribution on Jacobians of multivariate functions drawn from deep \gp{} priors, finding them to be products of independent Gaussian matrices with independent entries.

\begin{lemma}
\label{thm:deriv-ind}
The partial derivatives of a function mapping $\mathbb{R}^D \rightarrow \mathbb{R}$ drawn from a \gp{} prior with a product kernel are independently Gaussian distributed.
\end{lemma}
%
\begin{proof}
Because differentiation is a linear operator, the derivatives of a function drawn from a \gp{} prior are also jointly Gaussian distributed.
The covariance between partial derivatives with respect to input dimensions $d_1$ and $d_2$ of vector $\vx$ are given by \citet{Solak03derivativeobservations}:
\begin{align}
\cov \left( \frac{\partial f(\vx)}{\partial x_{d_1}}, \frac{\partial f(\vx)}{\partial x_{d_2}} \right) 
= \frac{\partial^2 k(\vx, \vx')}{\partial x_{d_1} \partial x_{d_2}'} \bigg|_{\vx=\vx'}
\label{eq:deriv-kernel}
\end{align}
If our kernel is a product over individual dimensions $k(\vx, \vx') = \prod_d^D k_d(x_d, x_d')$, 
then the off-diagonal entries are zero, implying that all elements are independent.
\end{proof}

For example, in the case of the multivariate squared-exp kernel, the covariance between derivatives has the form:
\begin{align}
& f(\vx) \sim \textnormal{GP}\left( 0, 
\sigma^2 \prod_{d=1}^D \exp \left(-\frac{1}{2} \frac{(x_d - x_d')^2}{\lengthscale_d^2} \right) \right) \nonumber \\
& \implies 
\cov \left( \frac{\partial f(\vx)}{\partial x_{d_1}}, \frac{\partial f(\vx)}{\partial x_{d_2}} \right) =
\begin{cases} 
\frac{\sigma^2}{\lengthscale_{d_1}^2} & \mbox{if } d_1 = d_2 \\ 
0 & \mbox{if } d_1 \neq d_2 \end{cases}
\end{align}

\begin{lemma}
\label{thm:matrix}
The Jacobian of a set of $D$ functions $\mathbb{R}^D \rightarrow \mathbb{R}$ drawn from independent \gp{} priors, each having product kernel is a $D \times D$ matrix of independent Gaussian R.V.'s
\end{lemma}
%
\begin{proof}
The Jacobian of the vector-valued function $\vf(\vx)$ is a matrix $J$ with elements ${J_{ij}(\vx) = \frac{ \partial f_i (\vx) }{\partial x_j}}$.
%
%
%
Because the \gp{}s on each output dimension $f_1(\vx), f_2(\vx), \dots, f_D(\vx)$ are independent by construction, it follows that each row of $J$ is independent.
Lemma \ref{thm:deriv-ind} shows that the elements of each row are independent Gaussian.
Thus all entries in the Jacobian of a \gp{}-distributed transform are independent Gaussian R.V.'s.
\end{proof}

\begin{theorem}
\label{thm:prodjacob}
The Jacobian of a deep \gp{} with a product kernel is a product of independent Gaussian matrices, with each entry in each matrix being drawn independently.
\end{theorem}
%
\begin{proof}
When composing $L$ different functions, we denote the \emph{immediate} Jacobian of the function mapping from layer $\layerindex -1$ to layer $\layerindex$ as $J^{(\layerindex)}(\vx)$, and the Jacobian of the entire composition of $L$ functions by $J^{(1:L)}(\vx)$.
By the multivariate chain rule, the Jacobian of a composition of functions is given by the product of the immediate Jacobian matrices of each function.  
Thus the Jacobian of the composed (deep) function $\vf^{(L)}(\vf^{(L-1)}(\dots \vf^{(3)}( \vf^{(2)}( \vf^{(1)}(\vx)))\dots))$ is
%
%
\begin{align}
 J^{(1:L)}(\vx) 
= J^{(L)} J^{(L-1)} \dots J^{(3)} J^{(2)} J^{(1)}.
\label{eq:jacobian-of-deep-gp}
\end{align}
By lemma \ref{thm:matrix}, each ${J^{(\layerindex)}_{i,j} \simind \mathcal{N}}$, so the complete Jacobian is a product of independent Gaussian matrices, with each entry of each matrix drawn independently.
\end{proof}

This result allows us to analyze the representational properties of a deep Gaussian process by examining the properties of products of independent Gaussian matrices.

\section{Formalizing a pathology}
\label{sec:formalizing-pathology}

\begin{figure}[h]
\centering
\begin{tikzpicture}[pile/.style={thick, ->, >=stealth'}]
    \node[anchor=south west,inner sep=0] at (0,0) {
    	\includegraphics[clip, trim = 0cm 12cm 0cm 0.0cm, width=0.6\columnwidth]{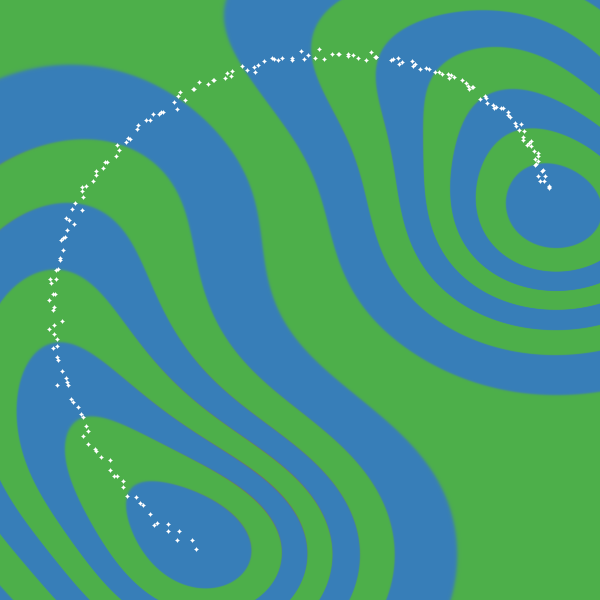}
    };
    \coordinate (D) at (1.6,1.5);
    \coordinate (Do) at (2.5, 0.8);
    \coordinate (Dt) at (3,3);
    
    \draw[pile] (D) -- (Dt) node[above, text width=5em] { tangent };
    \draw[pile] (D) -- (Do) node[right, text width=5em] { orthogonal };
\end{tikzpicture}
\caption[Desirable properties of representations of manifolds]
{Representing a 1-D data manifold.
Colors are a function of the computed representation of the input space.
The representation (blue \& green) changes little in directions orthogonal to the manifold (white), making it robust to noise in those directions.
The representation also varies in directions tangent to the data manifold, preserving information for later layers. 

}
\label{fig:hidden}
\end{figure}
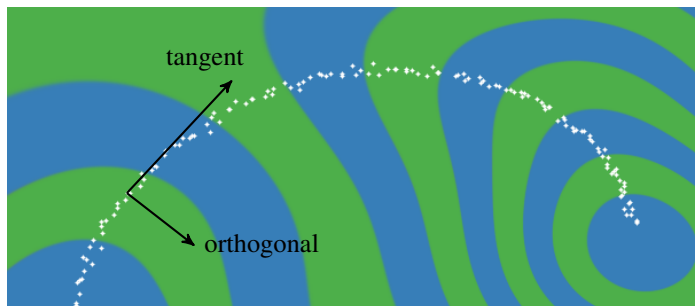

A common use of deep neural networks is building useful representations of data manifolds.
What properties make a representation useful?
\citet{rifai2011higher} argued that good representations of data manifolds are invariant in directions orthogonal to the data manifold.
They also argued that, conversely, a good representation must also change in directions tangent to the data manifold, in order to preserve relevant information.
\Cref{fig:hidden} visualizes a representation having these two properties.

As in \citet{rifai2011contractive}, we characterize the representational properties of a function by the singular value spectrum of the Jacobian.
The number of relatively large singular values of the Jacobian indicate the number of directions in data-space in which the representation varies significantly.
\newcommand{\spectrumpic}[1]{
\includegraphics[trim=4mm 1mm 4mm 2.5mm, clip, width=0.475\columnwidth]{\deeplimitsfiguresdir/spectrum/layer-#1}}%
\begin{figure}[h]
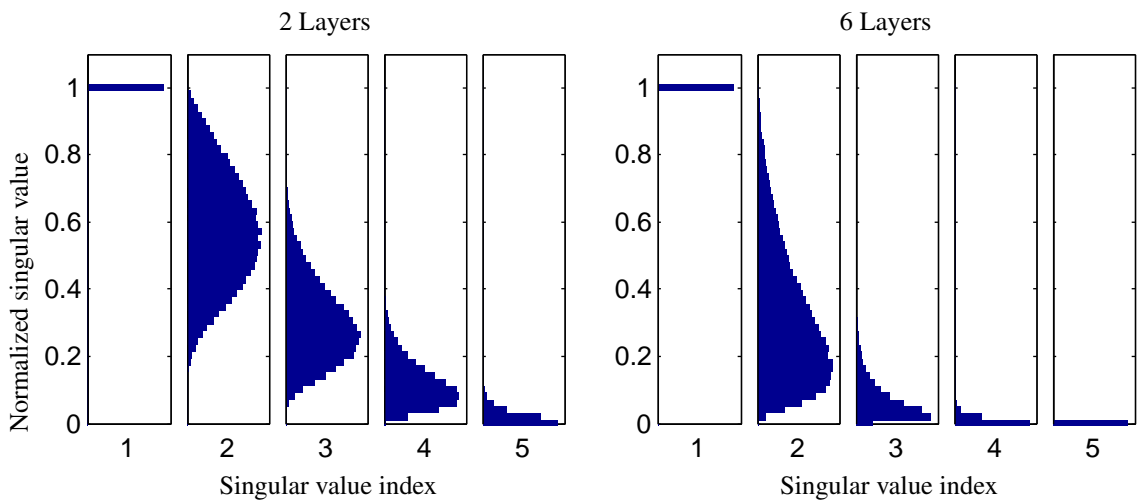

\centering
\begin{tabular}{ccc}
& 2 Layers & 6 Layers \\
\begin{sideways} { \quad Normalized singular value} \end{sideways} & \hspace{-0.2in} \spectrumpic{2} & \hspace{-0.1in} \spectrumpic{6} \\
 & { Singular value index} & { Singular value index}
\end{tabular}
\caption[Distribution of singular values of the Jacobian of a deep \sgp{}]
{
The distribution of normalized singular values of the Jacobian of a function drawn from a 5-dimensional deep \gp{} prior 2 layers deep~(\emph{Left}) and 6 layers deep~(\emph{Right}).
As nets get deeper, the largest singular value tends to become much larger than the others.
This implies that with high probability, these functions vary little in all directions but one, making them unsuitable for computing representations of manifolds of more than one dimension.
}
\label{fig:deep_spectrum}
\end{figure}%
\begin{figure}[h]%
\centering
\begin{tabular}{cc}
No transformation: $p(\vx)$ & 1 Layer: $p \left( \vf^{(1)}(\vx) \right)$ \\
\gpdrawbox{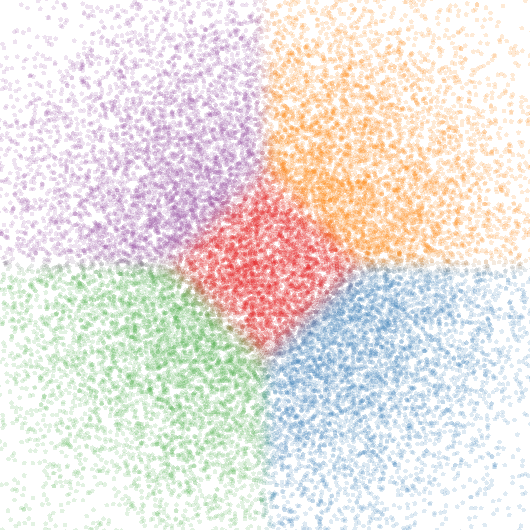} & \gpdrawbox{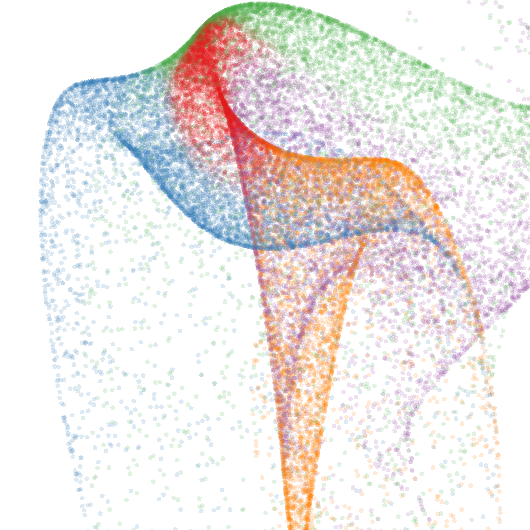} \\
4 Layers: $p \left( \vf^{(1:4)}(\vx) \right)$ & 6 Layers: $p \left( \vf^{(1:6)}(\vx) \right)$ \\
\gpdrawbox{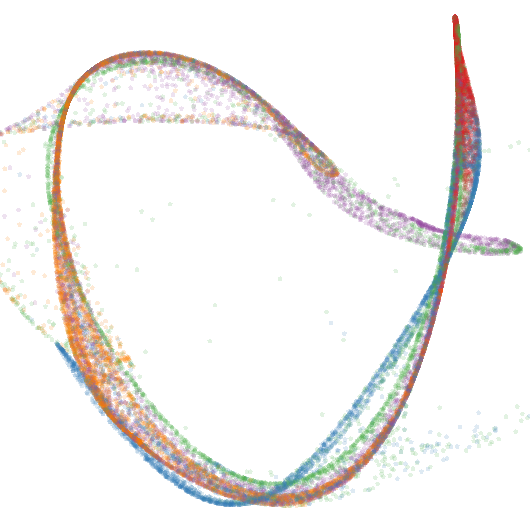} & \gpdrawbox{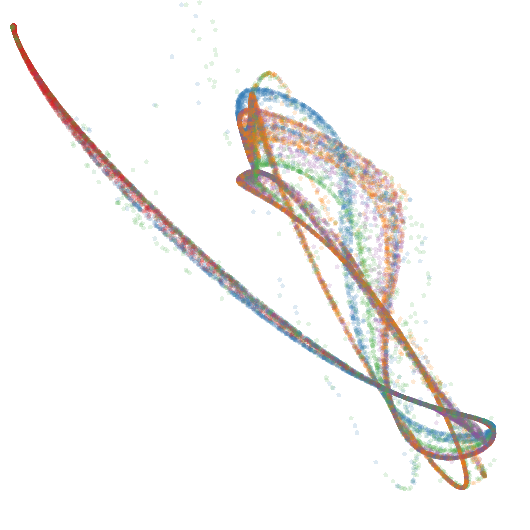}
\end{tabular}
\caption[Points warped by a draw from a deep \sgp{}]
{Points warped by a function drawn from a deep \gp{} prior.
\emph{Top left:} Points drawn from a 2-dimensional Gaussian distribution, color-coded by their location.
\emph{Subsequent panels:} Those same points, successively warped by compositions of functions drawn from a \gp{} prior.
As the number of layers increases, the density concentrates along one-dimensional filaments.
Warpings using random finite neural networks exhibit the same pathology, but also tend to concentrate density into 0-dimensional manifolds (points) due to saturation of all of the hidden units.}
\label{fig:filamentation}
\end{figure}%
\Cref{fig:deep_spectrum} shows the distribution of the singular value spectrum of draws from 5-dimensional deep \gp{}s of different depths.\footnote{\citet{rifai2011contractive} analyzed the Jacobian at location of the training points, but because the priors we are examining are stationary, the distribution of the Jacobian is identical everywhere.}
As the nets gets deeper, the largest singular value tends to dominate, implying there is usually only one effective degree of freedom in the representations being computed.

\Cref{fig:filamentation} demonstrates a related pathology that arises when composing functions to produce a deep density model.
The density in the observed space eventually becomes locally concentrated onto one-dimensional manifolds, or \emph{filaments}.
This again suggests that, when the width of the network is relatively small, deep compositions of independent functions are unsuitable for modeling manifolds whose underlying dimensionality is greater than one.

\newcommand{\mappic}[1]{ \includegraphics[width=0.475\columnwidth]{\deeplimitsfiguresdir/map/latent_coord_map_layer_#1} } 
\newcommand{\mappiccon}[1]{ \includegraphics[width=0.475\columnwidth]{\deeplimitsfiguresdir/map_connected/latent_coord_map_layer_#1} }
\begin{figure}
\centering
\begin{tabular}{cc}
\hspace{-0.15in} Identity Map: $\vy = \vx$ &
\hspace{-0.15in} 1 Layer: $\vy = \vf^{(1)}(\vx)$ \\
\hspace{-0.15in} \mappic{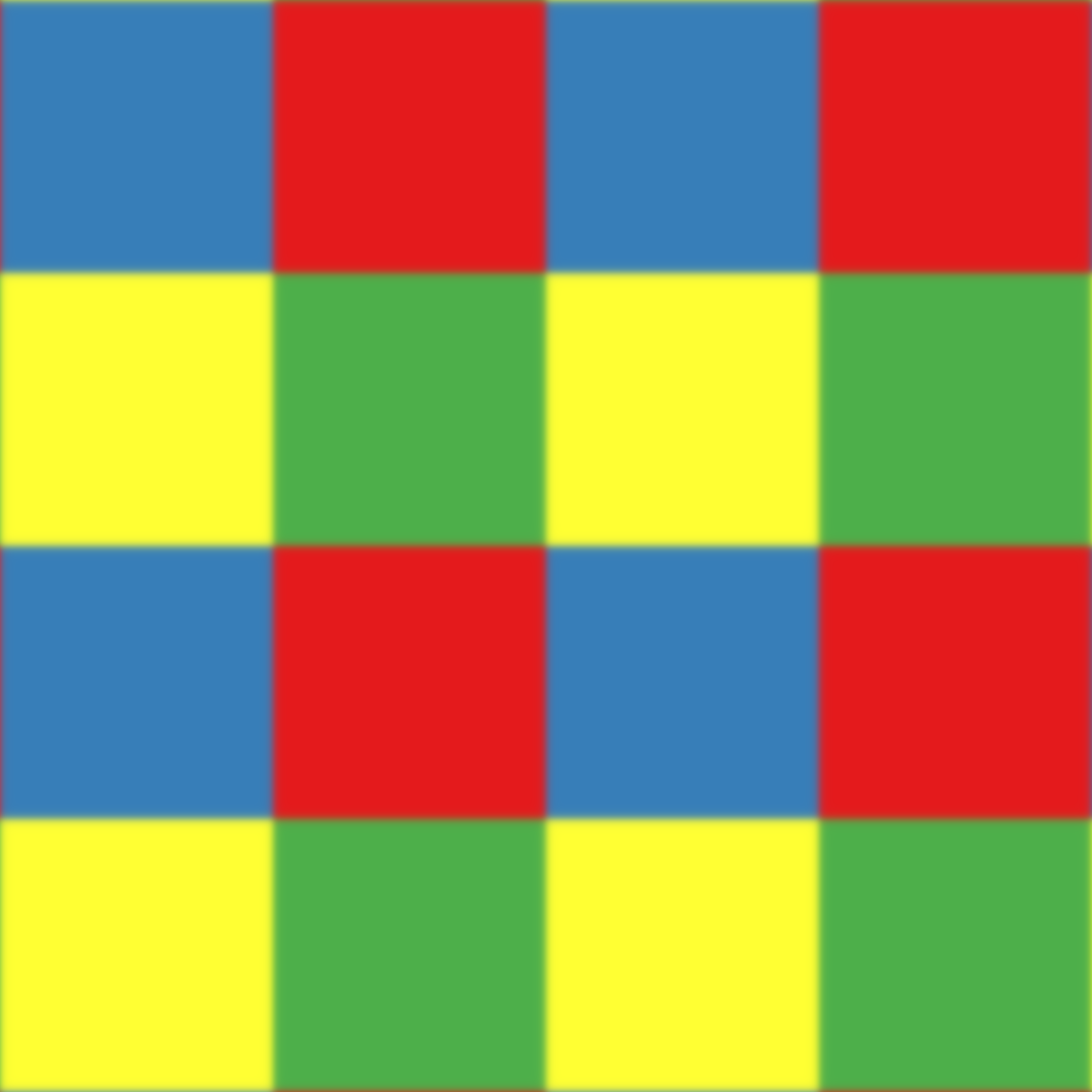} & \mappic{1} \\
\hspace{-0.15in} 10 Layers: $\vy = \vf^{(1:10)}(\vx)$ &
\hspace{-0.15in} 40 Layers: $\vy = \vf^{(1:40)}(\vx)$ \\
\hspace{-0.15in} \mappic{10} & \mappic{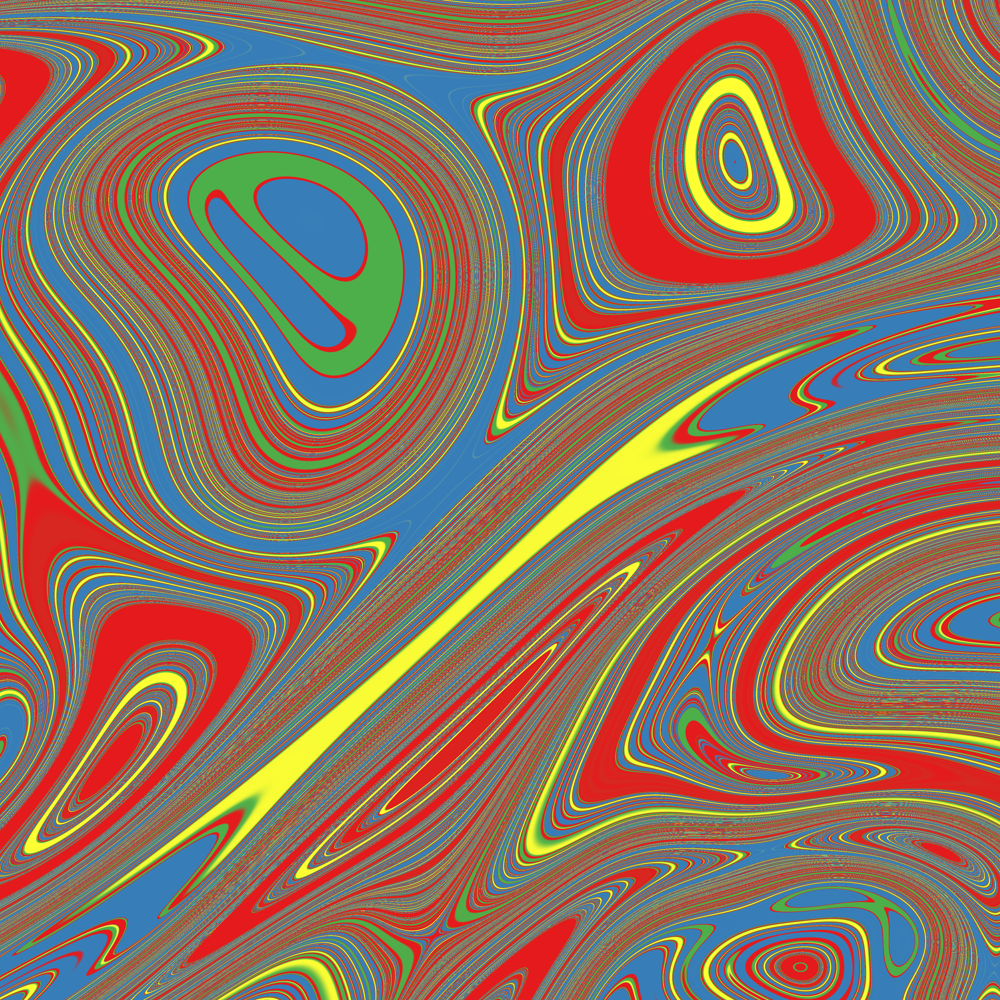}
\end{tabular}
\caption[Visualization of a feature map drawn from a deep \sgp{}]
{A visualization of the feature map implied by a function $\vf$ drawn from a deep \gp{}.
Colors are a function of the 2D representation $\vy = \vf(\vx)$ that each point is mapped to. 
The number of directions in which the color changes rapidly corresponds to the number of large singular values in the Jacobian.
Just as the densities in \cref{fig:filamentation} became locally one-dimensional, there is usually only one direction that one can move $\vx$ in locally to change $\vy$.
This means that $\vf$ is unlikely to be a suitable representation for decision tasks that depend on more than one aspect of $\vx$.  Also note that the overall shape of the mapping remains the same as the number of layers increase.
For example, a roughly circular shape remains in the top-left corner even after 40 independent warpings.}
\label{fig:deep_map}
\end{figure}
To visualize this pathology in another way, \cref{fig:deep_map} illustrates a color-coding of the representation computed by a deep \gp{}, evaluated at each point in the input space.
After 10 layers, we can see that locally, there is usually only one direction that one can move in $\vx$-space in order to change the value of the computed representation, or to cross a decision boundary.
This means that such representations are likely to be unsuitable for decision tasks that depend on more than one property of the input.

To what extent are these pathologies present in the types of neural networks commonly used in practice?
In simulations, we found that for deep functions with a fixed hidden dimension $D$, the singular value spectrum remained relatively flat for hundreds of layers as long as $D > 100$.
Thus, these pathologies are unlikely to severely effect the relatively shallow, wide networks most commonly used in practice.

\section{Fixing the pathology}
\label{sec:fix}

\begin{figure}[h]%
\def\nodeseptwo{1.9cm}%
\def\nodesize{.35cm}%
\def\numhiddentwo{3}%
\centering
\begin{tabular}{ccc}
a) Standard \MLP{} connectivity & &
b) Input-connected architecture\\
\hspace{-4mm}
\begin{tikzpicture}[draw=black!80]
    \tikzstyle{neuron}=[circle,minimum size=17pt, draw = black!80, fill = white, thick]
    \tikzstyle{input neuron}=[neuron, fill=green!50];
    \tikzstyle{output neuron}=[neuron, fill=red!50];
    \tikzstyle{hidden neuron}=[neuron, fill=blue!50];
    \tikzstyle{pile} =[thick, ->, >=stealth', shorten <=7pt, shorten >=8pt];

    \coordinate (I) at (0, 0);

    \foreach \name / \y in {1,...,\numhiddentwo} {
        \coordinate (H-\name) at (\nodeseptwo*\y, 0);
    }

    \path[pile] (I) edge (H-1) {};
    \foreach \name in {2,...,\numhiddentwo} {
	 	\pgfmathsetmacro\hindex{\name - 1}
		\path[pile] (H-\hindex) edge (H-\name) {};
    }

    \draw (I) node[neuron] {};
    \draw (I) node[below = 0.5cm]  {$\vx$};

    \foreach \name / \y in {1,...,\numhiddentwo} {
		\draw (H-\name) node[neuron]  {};
        \draw (H-\name) node[below = 0.34cm] {$\vf^{(\y)}(\vx)$};
    }
\end{tikzpicture} &
\hspace{0.5cm} &
\begin{tikzpicture}[draw=black!80]
    \tikzstyle{neuron}=[circle,minimum size=17pt, draw = black!80, fill = white, thick]
    \tikzstyle{input neuron}=[neuron, fill=green!50];
    \tikzstyle{output neuron}=[neuron, fill=red!50];
    \tikzstyle{hidden neuron}=[neuron, fill=blue!50];
    \tikzstyle{pile} =[thick, ->, >=stealth', shorten <=7pt, shorten >=8pt];

    \coordinate (I) at (0, 0);

    \foreach \name / \y in {1,...,\numhiddentwo} {
        \coordinate (H-\name) at (\nodeseptwo*\y, 0);
    }

    \path[pile] (I) edge (H-1) {};
    \foreach \name in {2,...,\numhiddentwo} {
		\pgfmathsetmacro\hindex{\name - 1}
		\path[pile] (H-\hindex) edge (H-\name) {};
        \path[pile] (I) edge [bend left] (H-\name) {};
    }

    \draw (I) node[neuron] {};
    \draw (I) node[below = 0.5cm]  {$\vx$};

    \foreach \name / \y in {1,...,\numhiddentwo} {
		\draw (H-\name) node[neuron]  {};
       	\draw (H-\name) node[below = 0.34cm] {$\vf_C^{(\y)}(\vx)$};
    }
\end{tikzpicture}
\end{tabular}
\caption[Two different architectures for deep neural networks]
{Two different architectures for deep neural networks.
\emph{Left:} The standard architecture connects each layer's outputs to the next layer's inputs.
\emph{Right:} The input-connected architecture also connects the original input $\vx$ to each layer.}
\label{fig:input-connected}
\end{figure}
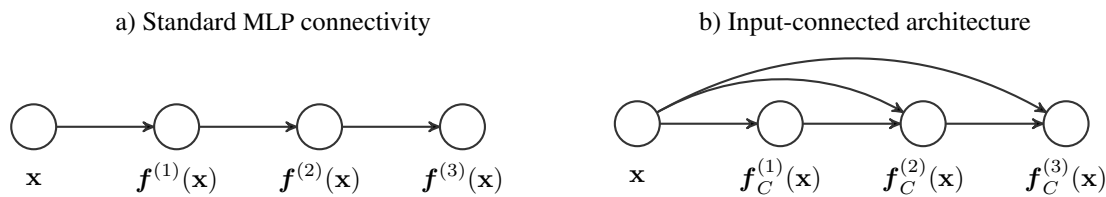

As suggested by \citet[chapter 2]{neal1995bayesian}, we can fix the pathologies exhibited in figures \cref{fig:filamentation} and \ref{fig:deep_map} by simply making each layer depend not only on the output of the previous layer, but also on the original input $\vx$.  
We refer to these models as \emph{input-connected} networks, and denote deep functions having this architecture with the subscript $C$, as in $f_C(\vx)$.
Formally, this functional dependence can be written as
\begin{align}
\vf_C^{(1:L)}(\vx) = \vf^{(L)} \left( \vf_C^{(1:L-1)}(\vx), \vx \right), \quad \forall L
\end{align}
\Cref{fig:input-connected} shows a graphical representation of the two connectivity architectures.

\begin{figure}[h]
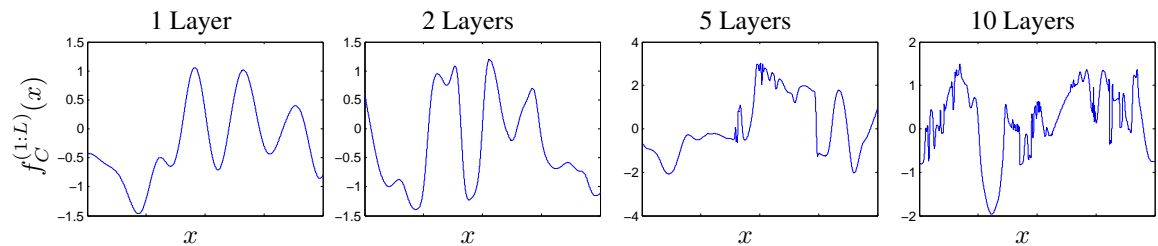

\centering
\setlength{\tabcolsep}{1.5pt}
\begin{tabular}{ccccc}
& 1 Layer & 2 Layers & 5 Layers & 10 Layers \\
\raisebox{0.6cm}{\rotatebox{90}{$f_C^{(1:L)}(x)$}} &
\onedsamplepiccon{1} &
\onedsamplepiccon{2} &
\onedsamplepiccon{5} &
\onedsamplepiccon{10} \\[-3pt]
 & $x$ & $x$ & $x$ & $x$
\end{tabular}
\caption[A draw from a 1D deep \sgp{} prior with each layer connected to the input]
{A draw from a 1D deep \gp{} prior having each layer also connected to the input.
The $x$-axis is the same for all plots.
Even after many layers, the functions remain relatively smooth in some regions, while varying rapidly in other regions.
Compare to standard-connectivity deep \gp{} draws shown in \cref{fig:deep_draw_1d}.}
\label{fig:deep_draw_1d_connected}
\end{figure}

\begin{figure}
\centering
\begin{tabular}{cc}
\hspace{-0.15in} Identity map: $\vy = \vx$ &
\hspace{-0.15in} 2 Connected layers: $\vy = \vf^{(1:2)}(\vx)$ \\
\hspace{-0.15in} \mappic{0} & \mappiccon{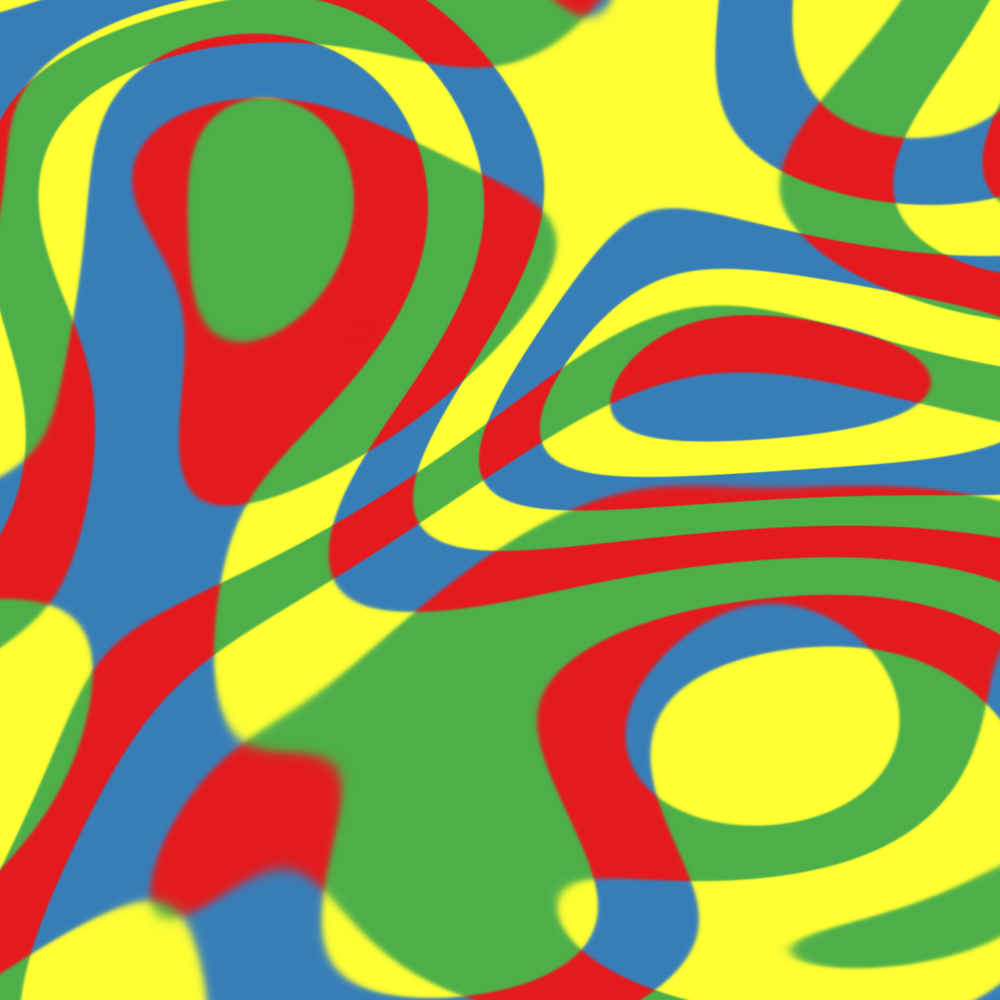} \\
\hspace{-0.15in} 10 Connected layers: $\vy = \vf^{(1:10)}(\vx)$ &
\hspace{-0.15in} 20 Connected layers: $\vy = \vf^{(1:20)}(\vx)$ \\
\hspace{-0.15in} \mappiccon{10} & \mappiccon{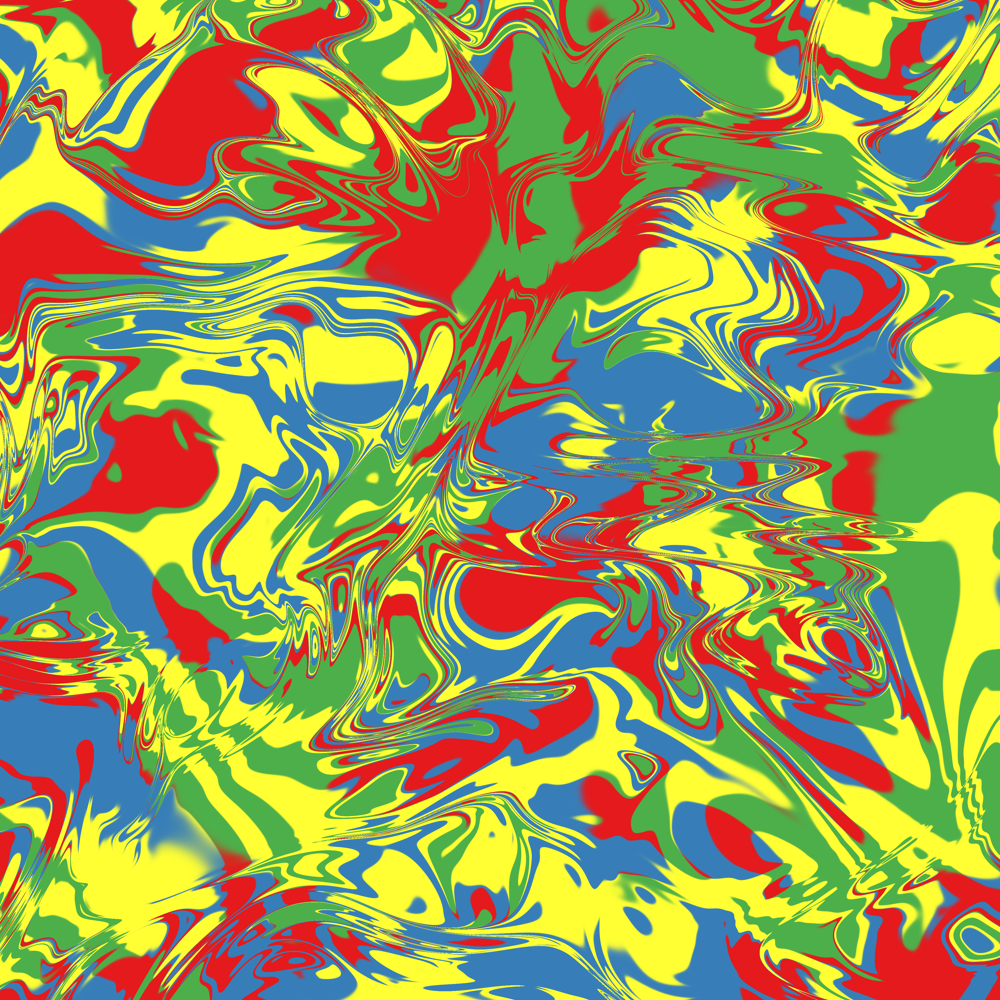}
\end{tabular}
\caption[Feature map of an input-connected deep \sgp{}]
{The feature map implied by a function $\vf$ drawn from a deep \gp{} prior with each layer also connected to the input $\vx$, visualized at various depths.
Compare to the map shown in \cref{fig:deep_map}.
In the mapping shown here there are sometimes two directions that one can move locally in $\vx$ to in order to change the value of $\vf(\vx)$.
This means that the input-connected prior puts significant probability mass on a greater variety of types of representations, some of which depend on all aspects of the input.
}
\label{fig:deep_map_connected}
\end{figure}

Similar connections between non-adjacent layers can also be found the primate visual cortex \citep{maunsell1983connections}.
Visualizations of the resulting prior on functions are shown in \cref{fig:deep_draw_1d_connected,fig:no_filamentation,fig:deep_map_connected}.

\newcommand{\gpdrawboxcon}[1]{
\setlength\fboxsep{0pt}
\hspace{-0.2in} 
\fbox{
\includegraphics[width=0.464\columnwidth]{\deeplimitsfiguresdir/deep_draws_connected/deep_sample_connected_layer#1}
}}
\begin{figure}
\centering
\begin{tabular}{cc}
3 Connected layers: $p \left( \vf_C^{(1:3)}(\vx) \right)$ & 6 Connected layers: $p \left( \vf_C^{(1:6)}(\vx) \right)$ \\
\gpdrawboxcon{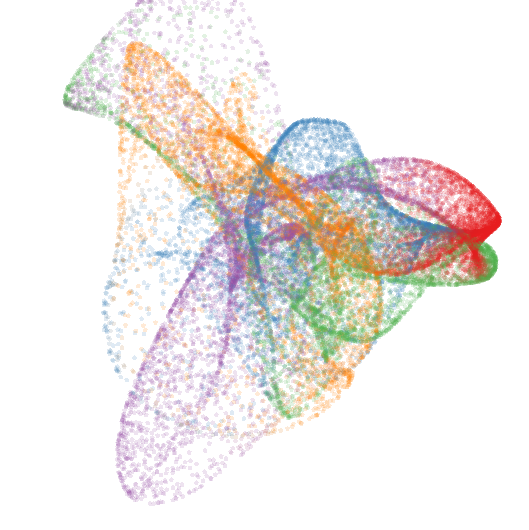} &
\gpdrawboxcon{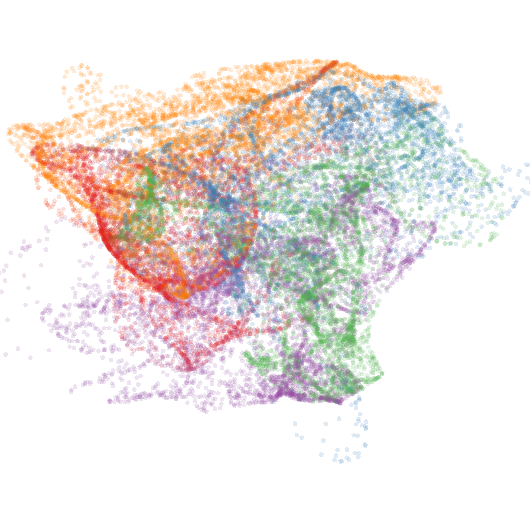}
\end{tabular}
\caption[Points warped by a draw from an input-connected deep \sgp{}]
{Points warped by a draw from a deep \sgp{} with each layer connected to the input $\vx$.
As depth increases, the density becomes more complex without concentrating only along one-dimensional filaments.}
\label{fig:no_filamentation}
\end{figure}
\begin{figure}
\centering
\newcommand{\spectrumpiccon}[1]{
\includegraphics[trim=4mm 1mm 4mm 2.5mm, clip, width=0.475\columnwidth]{\deeplimitsfiguresdir/spectrum/con-layer-#1}} 
\begin{tabular}{ccc}
 & 25 layers &  50 layers \\
\hspace{-0.5cm} \begin{sideways} {\quad Normalized singular value} \end{sideways} & \hspace{-0.2in} \spectrumpiccon{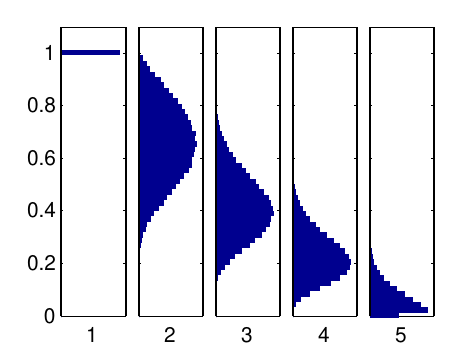} & \hspace{-0.16in} \spectrumpiccon{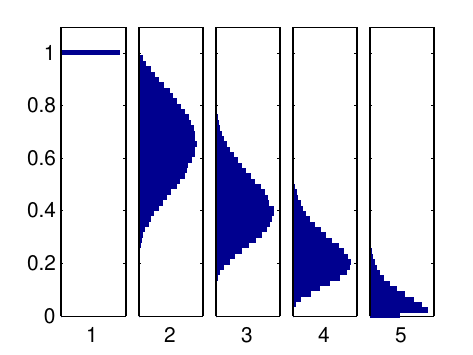} \\
 & {Singular value number} & {Singular value number}
\end{tabular}
\caption[Distribution of singular values of an input-connected deep \sgp{}]
{The distribution of singular values drawn from 5-dimensional input-connected deep \gp{} priors, 25 layers deep (\emph{Left}) and 50 layers deep (\emph{Right}).
Compared to the standard architecture, the singular values are more likely to remain the same size as one another, meaning that the model outputs are more often sensitive to several directions of variation in the input.}
\label{fig:good_spectrum}
\end{figure}

The Jacobian of an input-connected deep function is defined by the recurrence
\newcommand{\sbi}[2]{\left[ \! \begin{array}{c} #1 \\ #2 \end{array} \! \right]} 
\begin{align}
{J_C^{(1:L)} = J^{(L)} \sbi{ J_C^{(1:L-1)}}{I_D}}.
\end{align}
%
%
%
where $I_D$ is a $D$-dimensional identity matrix.
Thus the Jacobian of an input-connected deep \gp{} is a product of independent Gaussian matrices each with an identity matrix appended.
\Cref{fig:good_spectrum} shows that with this architecture, even 50-layer deep \gp{}s have well-behaved singular value spectra.

The pathology examined in this section is an example of the sort of analysis made possible by a well-defined prior on functions.
The figures and analysis done in this section could be done using Bayesian neural networks with finite numbers of nodes, but would be more difficult.
In particular, care would need to be taken to ensure that the networks do not produce degenerate mappings due to saturation of the hidden units.

\section{Deep kernels}
\label{sec:deep_kernels}


\cite{ bengio2006curse} showed that kernel machines have limited generalization ability when they use ``local'' kernels such as the squared-exp.
However, many interesting non-local kernels can be constructed which allow some forms of extrapolation.
One way to build non-local kernels is by composing fixed feature maps, creating \emph{deep kernels}.
For example, periodic kernels can be viewed as a 2-layer-deep kernel, in which the first layer maps $x \rightarrow [\sin(x), \cos(x)]$, and the second layer maps through basis functions corresponding to the implicitly feature map giving rise to an \kSE{} kernel.

This section builds on the work of \citet{cho2009kernel}, who derived several kinds of deep kernels by composing multiple layers of feature mappings.

 

In principle, one can compose the implicit feature maps of any two kernels $k_a$ and $k_b$ to get a new kernel, which we denote by $\left( k_b \circ k_a \right)$:
\begin{align}
k_a(\vx, \vx') & = \hPhi_a(\vx) \tra \hPhi_a(\vx') \\
k_b(\vx, \vx') & = \hPhi_b(\vx) \tra \hPhi_b(\vx') \\
\left( k_b \circ k_a \right)(\vx, \vx') & = k_b \big(\hPhi_a(\vx), \hPhi_a(\vx') \big) 
 = \left[ \hPhi_b \left( \hPhi_a(\vx) \right)\right] \tra \hPhi_b \left(\hPhi_a(\vx') \right)
\end{align}
However, this composition might not always have a closed form if the number of hidden features of either kernel is infinite.

Fortunately, composing the squared-exp (\kSE{}) kernel with the implicit mapping given by any other kernel has a simple closed form.
If $k(\vx, \vx') = \hPhi(\vx)\tra \hPhi(\vx')$, then
%
\begin{align}
\left( \kSE \circ k \right) \left( \vx, \vx' \right) & = k_{SE} \big( \hPhi(\vx), \hPhi(\vx') \big) \\
& = \exp \left( -\frac{1}{2} || \hPhi(\vx) - \hPhi(\vx')||_2^2 \right) \\
& = \exp\left ( -\frac{1}{2} \left[ \hPhi(\vx) \tra \hPhi(\vx) - 2 \hPhi(\vx) \tra \hPhi(\vx') + \hPhi(\vx') \tra \hPhi(\vx') \right] \right)  \\
& = \exp \left( -\frac{1}{2} \big[ k(\vx, \vx) - 2 k(\vx, \vx') + k(\vx', \vx') \big] \right) \; .
\end{align}
%
%
%
This formula expresses the composed kernel $(\kSE \circ k)$ exactly in terms of evaluations of the original kernel $k$.

\subsection{Infinitely deep kernels}
What happens when one repeatedly composes feature maps many times, starting with the squared-exp kernel?
If the output variance of the \kSE{} is normalized to ${k(\vx,\vx) = 1}$, then the infinite limit of composition with \kSE{} converges to ${\left(\kSE \circ \kSE \circ \ldots \circ \kSE\right)(\vx,\vx') = 1}$ for all pairs of inputs.
A constant covariance corresponds to a prior on constant functions ${f(\vx) = c}$.
This can be viewed as a degenerate limit.

%


As above, we can overcome this degeneracy by connecting the input $\vx$ to each layer.
To do so, we concatenate the composed feature vector at each layer, $\hPhi^{(1:\ell)}(\vx)$, with the input vector $\vx$ to produce an input-connected deep kernel $k_C^{(1:L)}$, defined by:
\begin{align}
k_C^{(1:\ell + 1)}(\vx, \vx')
& = \exp \left( -\frac{1}{2} \left|\left| 
\left[ \! \begin{array}{c} \hPhi^{(1:\ell)}(\vx) \\ \vx \end{array} \! \right] - 
\left[ \! \begin{array}{c} \hPhi^{(1:\ell)}(\vx') \\ \vx' \end{array} \! \right] \right| \right|_2^2 \right) \\
& = \exp \Big( -\frac{1}{2} \big[ k_C^{(1:\ell)}(\vx, \vx) - 2 k_C^{(1:\ell)}(\vx, \vx') 
 + k_C^{(1:\ell)}(\vx', \vx') {\color{black} - || \vx - \vx' ||_2^2} \big] \Big) \nonumber
\end{align}
Starting with the squared-exp kernel, this repeated mapping satisfies
\begin{align}
k_C^{(1:\infty)}(\vx, \vx') - \log \left( k_C^{(1:\infty)}(\vx, \vx') \right) = 1 + \frac{1}{2} || \vx - \vx' ||_2^2 \,.
\end{align}
\begin{figure}
\centering
\begin{tabular}{cc}
Input-connected deep kernels & Draws from corresponding \gp{}s \\
\hspace{-0.3cm}
\rotatebox{90}{$\qquad \qquad k(x, x')$}
\includegraphics[width=0.465\columnwidth, clip, trim = 1.3cm 0.4cm 0.9cm 0.3cm]{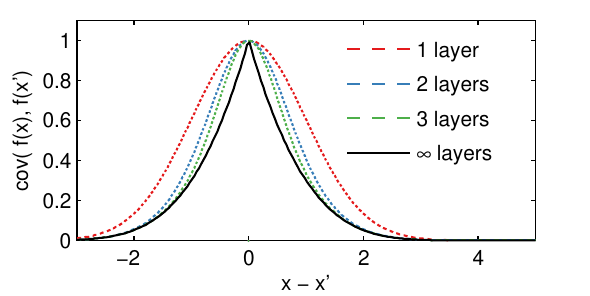} &
\hspace{-0.3cm}
\rotatebox{90}{$\qquad \qquad f(x)$}
\includegraphics[width=0.44\columnwidth, clip, trim = 1.29cm 0.1cm 0.9cm 0.35cm]{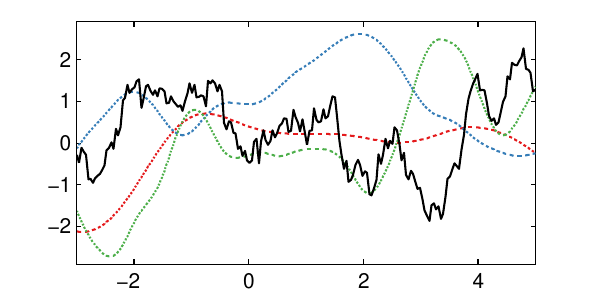} \\
$x - x'$ &  $x$
\end{tabular}
\caption[Infinitely deep kernels]{
\emph{Left:} Input-connected deep kernels of different depths.
By connecting the input $\vx$ to each layer, the kernel can still depend on its input even after arbitrarily many layers of composition.
\emph{Right:} Draws from \gp{}s with deep input-connected kernels.}
\label{fig:deep_kernel_connected}
\end{figure}
The solution to this recurrence is related to the Lambert W-function~\citep{corless1996lambertw} and has no closed form.
In one input dimension, it has a similar shape to the Ornstein-Uhlenbeck covariance ${\OU(x,x') = \exp( -|x - x'| )}$ but with lighter tails.
Samples from a \gp{} prior having this kernel are not differentiable, and are locally fractal.
\Cref{fig:deep_kernel_connected} shows this kernel at different depths, as well as samples from the corresponding \gp{} priors.

One can also consider two related connectivity architectures: one in which each layer is connected to the output layer, and another in which every layer is connected to all subsequent layers.
It is easy to show that in the limit of infinite depth of composing \kSE{} kernels, both these architectures converge to $k(\vx, \vx') = \delta( \vx, \vx' )$, the white noise kernel.

\subsection{When are deep kernels useful models?}

Kernels correspond to fixed feature maps, and so kernel learning is an example of implicit representation learning. 
Kernels can capture complex structure \citep{DuvLloGroetal13}, and can enable many types of generalization, such as translation and rotation invariance in images \citep{kondor2008group}.
More generally, \cite{salakhutdinov2008using} used a deep neural network to learn feature transforms for kernels, to learn invariances in an unsupervised manner.
We believe that the relatively uninteresting properties of the deep kernels derived in this section simply reflect the fact that an arbitrary computation, even if it is ``deep'', is not likely to give rise to a useful representation unless combined with learning.
To put it another way, any fixed representation is unlikely to be useful unless it has been chosen specifically for the problem at hand.

\section{Dropout in Gaussian processes}
\label{sec:dropout-gps}

\emph{Dropout} is a recently-introduced method for regularizing neural networks~\citep{hinton2012improving, srivastava2013improving}.
Training with dropout entails independently setting to zero (``dropping'') some proportion $p$ of features or inputs, in order to improve the robustness of the resulting network, by reducing co-dependence between neurons.
To maintain similar overall activation levels, the remaining weights are divided by $p$.
Predictions are made by approximately averaging over all possible ways of dropping out neurons.

\citet{baldi2013understanding} and \citet{wang2013fast} analyzed dropout in terms of the effective prior induced by this procedure in several models, such as linear and logistic regression.
In this section, we perform a similar analysis for \gp{}s, examining the priors on functions that result from performing dropout in the one-hidden-layer neural network implicitly defined by a \gp{}.

Recall from \cref{sec:relating} that some \gp{}s can be derived as infinitely-wide one-hidden-layer neural networks, with fixed activation functions $\feat(\vx)$ and independent random weights $\vw$ having zero mean and finite variance $\sigma^2_{\vw}$:
\begin{align}
f(\vx) 
= \frac{1}{K} \sum_{i=1}^K \netweights_i \hphi_i(\vx)
\implies f \distas{K \to \infty} \GPt{0}{\sigma^2_{\vw} \feat(\vx)\tra\feat(\vx')} .
\label{eq:one-layer-gp-two}
\end{align}
%

\subsection{Dropout on infinitely-wide hidden layers has no effect}

First, we examine the prior obtained by dropping features from $\hPhi(\vx)$ by setting weights in $\vnetweights$ to zero independently with probability $p$.
For simplicity, we assume that $\expectargs{}{\vnetweights} = \vzero$.
If the weights $w_i$ initially have finite variance $\sigma^2_{\netweights}$ before dropout, then the weights after dropout (denoted by $r_i w_i$, where $r_i$ is a Bernoulli random variable) will have variance:
\begin{align}
r_i \simiid \textnormal{Ber}(p) \qquad
\varianceargs{}{r_i \netweights_i} = p \sigma_{\netweights}^2 \;.
\end{align}
Because \cref{eq:one-layer-gp-two} is a result of the central limit theorem, it does not depend on the exact form of the distribution on $\vnetweights$, but only on its mean and variance.
Thus the central limit theorem still applies.
Performing dropout on the features of an infinitely-wide \MLP{} does not change the resulting model at all, except to rescale the output variance.
Indeed, dividing all weights by $\sqrt p$ restores the initial variance:
\begin{align}
\varianceargs{}{ \frac{1}{\sqrt p} r_i \netweights_i} = \frac{p}{p} \sigma_{\netweights}^2 = \sigma_{\netweights}^2
\end{align}
in which case dropout on the hidden units has no effect at all.
Intuitively, this is because no individual feature can have more than an infinitesimal contribution to the network output.

This result does not hold in neural networks having a finite number of hidden features with Gaussian-distributed weights, another model class that also gives rise to \gp{}s.

\subsection{Dropout on inputs gives additive covariance}
One can also perform dropout on the $D$ inputs to the \gp{}.
For simplicity, consider a stationary product kernel ${k(\vx, \vx') = \prod_{d=1}^D k_d(x_d, x_d')}$ which has been normalized such that $k(\vx, \vx) = 1$, and a dropout probability of $p = \nicefrac{1}{2}$.
In this case, the generative model can be written as:
\begin{align}
\vr = [r_1, r_2, \dots, r_D], \quad \textnormal{each} \;\; r_i \simiid \textnormal{Ber} \left( \frac{1}{2} \right), \quad f(\vx) | \vr \sim \GP \left( 0, \prod_{d=1}^D k_d(x_d, x_d')^{r_d} \right)
\end{align}
This is a mixture of $2^D$ \gp{}s, each depending on a different subset of the inputs:
\begin{align}
p \left( f(\vx) \right) = 
\sum_{\vr} p \left( f(\vx) | \vr \right) p( \vr) = 
\frac{1}{2^D} \sum_{\vr \in \{0,1\}^D}  \GP \left(f(\vx) \,\Big|\, 0, \prod_{d=1}^D k_d(x_d, x_d')^{r_d} \right)
\label{eq:dropout-mixture}
\end{align}
We present two results which might give intuition about this model.

\paragraph{Interpretation as a spike-and-slab prior on lengthscales}
First, if the kernel on each dimension has the form ${k_d(x_d, x_d') = g \left( \frac{x_d - x_d'}{\ell_d} \right)}$, as does the \kSE{} kernel, then any input dimension can be dropped out by setting its lengthscale $\ell_d$ to $\infty$.
In this case, performing dropout on the inputs of a \gp{} corresponds to putting independent spike-and-slab priors on the lengthscales, with each dimension's distribution independently having ``spikes'' at $\ell_d = \infty$ with probability mass of $\nicefrac{1}{2}$.

\paragraph{Interpretation as an additive GP}
Another way to understand the resulting prior is to note that the dropout mixture given by \cref{eq:dropout-mixture} has the following covariance:
\begin{align}
\cov\left( \colvec{f(\vx)}{f(\vx')} \right) = \frac{1}{2^{D}} \sum_{\vr \in \{0,1\}^D}  \prod_{d=1}^D k_d(x_d, x_d')^{r_d}
\label{eq:dropout-mixture-covariance}
\end{align}
For dropout rates $p \neq \nicefrac{1}{2}$, the $d$th order terms will be weighted by $p^{(D - d)}(1-p)^d$.

\begin{figure}[h!]
\centering
\renewcommand{\tabcolsep}{0pt}
\begin{tabular}{cccc}
1st-order terms: & 2nd-order terms: &  3rd-order terms: & Sum of all terms: \\
$k_1 + k_2 + k_3$ & $k_1k_2 + k_2k_3 + k_1k_3$ & $k_1k_2k_3$ ($\kSE$ kernel) & Dropout kernel \\
\includegraphics[trim=1em 0em 1em 3em, clip, width=0.25\textwidth]{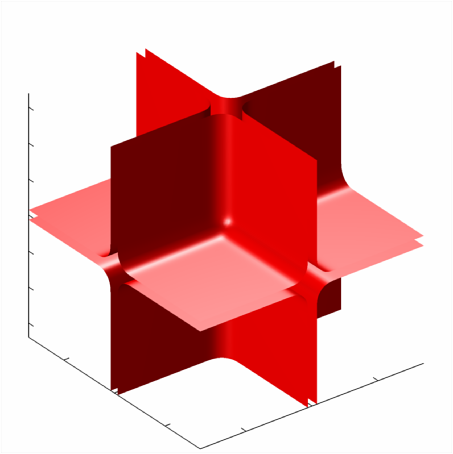} &
\includegraphics[trim=1em 0em 1em 3em, clip, width=0.25\textwidth]{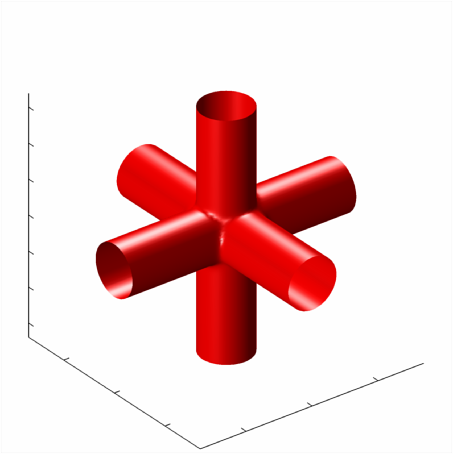} &
\includegraphics[trim=1em 0em 1em 3em, clip, width=0.25\textwidth]{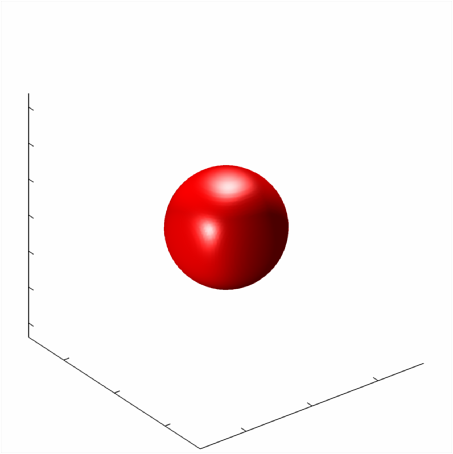} & 
\includegraphics[trim=1em 0em 1em 3em, clip, width=0.25\textwidth]{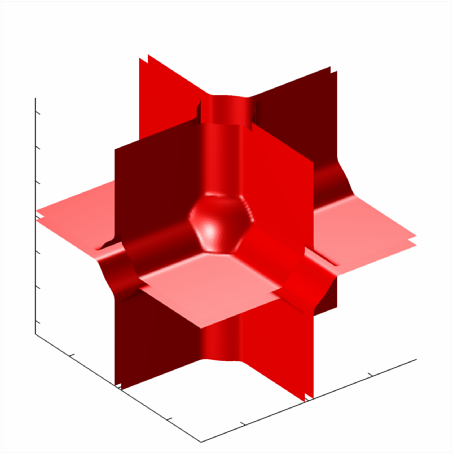}\\
$\vx - \vx'$ & $\vx - \vx'$ & $\vx - \vx'$ & $\vx - \vx'$\\[0.5em]
\end{tabular}
\caption[Isocontours of additive kernels in 3 dimensions]
{Isocontours of dropout kernel in $D = 3$ dimensions.
The $D$th-order kernel only considers nearby points relevant, while lower-order kernels allow the output to depend on distant training points, as long as they share one or more input value with a test point.
Figure taken from \citet{duvenaud2011additive11}.
}
\label{fig:kernels3d}
\end{figure}

Therefore, performing dropout on the inputs of a \gp{} gives a non-Gaussian distribution that has the same first two moments as a \gp{} having a covariance given by \cref{eq:dropout-mixture-covariance}.
This model class is called \emph{additive} \gp{}s, and have the property that they can sometimes allow non-local extrapolation~\citep{duvenaud2011additive11}.
To see why, note that a \gp{} whose covariance is a sum of kernels corresponds to a sum of functions, each distributed according to a \gp{} having the corresponding kernel.
Therefore, \cref{eq:dropout-mixture-covariance} describes a prior on a sum of $2^D$ functions, each depending on a different subset of input variables.
Functions which depend only on a small number of input dimensions will be invariant to noise in all dimensions on which they don't depend.
Isocontours of the resulting kernels are shown in \Cref{fig:kernels3d}.

Thus intepreting dropout as a mixture of models, in which each model only depends on a subset of the input variables, helps to explain why dropout sometimes leads to better generalization.

\section{Related work}

\subsubsection{Deep Gaussian processes}
\citet[chapter 2]{neal1995bayesian} explored properties of arbitrarily deep Bayesian neural networks, including those that would give rise to deep \gp{}s.
He noted that infinitely deep random neural networks without extra connections to the input would be equivalent to a Markov chain, and therefore would lead to degenerate priors on functions.
He also suggested connecting the input to each layer in order to fix this problem.
Much of the analysis in this paper can be seen as a more detailed investigation, and vindication, of these claims.

The first instance of deep \gp{}s being used in practice was \citep{lawrence2007hierarchical}, who presented a model called ``hierarchical \gplvm{}s'', in which time was mapped through a composition of multiple \gp{}s to produce observations.

The term ``deep Gaussian processes'' was first used by \citet{damianou2012deep}, who developed a variational inference method, analyzed the effect of automatic relevance determination, and showed that deep \gp{}S could learn with relatively little data.
%
They used the term ``deep \gp{}'' to refer both to supervised models (compositions of \gp{}s) and to unsupervised models (compositions of \gplvm{}s).
This conflation may be reasonable, since the activations of the hidden layers are themselves latent variables, even in supervised settings:
Depending on kernel parameters, each latent variable may or may not depend on the layer below.

In general, supervised models can also be latent-variable models.
For example, \citet{wang2012gaussian} investigated single-layer \gp{} regression models that had additional latent inputs.

\subsubsection{Nonparametric neural networks}
\citet{adams2010learning} proposed a prior on arbitrarily deep Bayesian networks having an unknown and unbounded number of parametric hidden units in each layer.
Their architecture has connections only between adjacent layers, and so may have similar pathologies to the one discussed in this paper as the number of layers increases.

\citet{wilson2012gaussian} introduced Gaussian process regression networks, which are defined as a matrix product of draws from \gp{}s priors, rather than a composition.
These networks have the form:
%
\begin{align}
\vy(\vx) & = \vW(\vx) \vf(\vx) \qquad \textnormal{where each} \; f_{d}, W_{d,j} \simiid \GPt{\vzero}{\kSE} + \Nt{0}{\sigma^2_n} .
\end{align}
We can easily define a ``deep'' Gaussian process regression network:
\begin{align}
\vy(\vx) = \vW^{(3)}(\vx) \vW^{(2)}(\vx) \vW^{(1)}(\vx) \vf(\vx)
\end{align}
which repeatedly adds and multiplies functions drawn from \gp{}s, in contrast to deep \gp{}s, which repeatedly compose functions.
This prior on functions has a similar form to the Jacobian of a deep \gp{} (\cref{eq:jacobian-of-deep-gp}), and so might be amenable to a similar analysis to that of section \ref{sec:characterizing-deep-gps}.

%
%
%

\subsubsection{Information-preserving architectures}
Deep density networks \citep{rippel2013high} are constructed through a series of parametric warpings of fixed dimension, with penalty terms encouraging the preservation of information about lower layers.
This is another promising approach to fixing the pathology discussed in \cref{sec:formalizing-pathology}.

\subsubsection{Recurrent networks}
\citet{bengio1994learning} and \citet{pascanu2012understanding} analyzed a related problem with gradient-based learning in recurrent networks, the ``exploding-gradients'' problem.
They noted that in recurrent neural networks, the size of the training gradient can grow or shrink exponentially as it is back-propagated, making gradient-based training difficult.

\citet{hochreiter1997long} addressed the exploding-gradients problem by introducing hidden units designed to have stable gradients.
This architecture is known as long short-term memory.

\subsubsection{Deep kernels}

The first systematic examination of deep kernels was done by \citet{cho2009kernel}, who derived closed-form composition rules for $\kSE$, polynomial, and arc-cosine kernels, and showed that deep arc-cosine kernels performed competitively in machine-vision applications when used in a \SVM{}.

\citet{hermans2012recurrent} constructed deep kernels in a time-series setting, constructing kernels corresponding to infinite-width \emph{recurrent} neural networks.
They also proposed concatenating the implicit feature vectors from previous time-steps with the current inputs, resulting in an architecture analogous to the input-connected architecture proposed by \citet[chapter 2]{neal1995bayesian}.

\subsubsection{Analyses of deep learning}
\citet{montavon2010layer} performed a layer-wise analysis of deep networks, and noted that the performance of \MLP{}s degrades as the number of layers with random weights increases, a result consistent with the analysis of this paper.

The experiments of \citet{saxe2011random} suggested that most of the good performance of convolutional neural networks could be attributed to the architecture alone.
Later, \citet{saxedynamics} looked at the dynamics of gradient-based training methods in deep \emph{linear} networks as a tractable approximation to standard deep (nonlinear) neural networks.



\subsubsection{Source code}
Source code to produce all figures is available at \url{http://www.github.com/duvenaud/deep-limits}.
This code is also capable of producing visualizations of mappings such as \cref{fig:deep_map,fig:deep_map_connected} using neural nets instead of \gp{}s at each layer.

\section{Conclusions}


This paper used well-defined priors to explicitly examine the assumptions made by neural network models.
%
We used deep Gaussian processes as an easy-to-analyze model corresponding to multi-layer preceptrons having nonparametric activation functions.
%
We showed that representations based on repeated composition of independent functions exhibit a pathology where the representations becomes invariant to all but one direction of variation. 
We then showed that this problem could be alleviated by connecting the input to each layer.

We also examined properties of deep kernels, corresponding to arbitrarily many compositions of fixed feature maps.
Finally, we derived models obtained by performing dropout on Gaussian processes, finding a tractable approximation to exact dropout in \gp{}s.

Much recent work on deep networks has focused on weight initialization \citep{martens2010deep}, regularization \citep{lee2007sparse} and network architecture \citep{gens2013learning}.
If we can identify priors that give our models desirable properties, these might in turn suggest regularization, initialization, and architecture choices that also provide such properties.


\subsubsection*{Acknowledgements}
We thank Carl Rasmussen, Andrew McHutchon, Neil Lawrence, Andreas Damianou, James Robert Lloyd, Creighton Heaukulani, Dan Roy, Mark van der Wilk, Miguel Hern\'{a}ndez-Lobato and Andrew Tulloch for helpful discussions.

\bibliographystyle{plainnat}
\bibliography{references.bib}

\end{document}

%% file: preamble.tex
\newcommand{\deeplimitsfiguresdir}{figures/deep-limits}

\newcommand{\additivefigsdir}{figures/additive}

\usepackage{bm}  
\usepackage{booktabs}
\usepackage{tabularx}
\usepackage{multirow}
\usepackage{datetime}

\usepackage{relsize}
\usepackage{graphicx}
\usepackage{amsmath,amssymb,textcomp}
\usepackage{nicefrac}
\usepackage{amsthm}
\usepackage{tikz}
\usetikzlibrary{arrows}
\usetikzlibrary{calc}
\usepackage{nth}
\usepackage{rotating}
\usepackage{array}
\usepackage{fp}
\usepackage{cleveref}   
\crefname{equation}{equation}{equations}
\crefname{figure}{figure}{figures}

\definecolor{mydarkblue}{rgb}{0.00,0.08,0.45}
\usepackage{url}
    \hypersetup{
      plainpages=false,
      pdfstartview=FitV, 
      pdftoolbar=true, 
      pdfmenubar=true,
      bookmarksopen=true,
      bookmarksnumbered=true,
      breaklinks=true,
      linktocpage, 
      colorlinks=true, 
      anchorcolor=black,
      linkcolor=mydarkblue, 
      urlcolor=mydarkblue,
      citecolor=mydarkblue, 
    }
    
\usepackage[hyperpageref]{backref}
\newcommand{\comfyfill}[1]{
  \unskip\hspace*{0.1em plus 1fill}
  \nolinebreak[3]%
  \hspace*{\fill}\mbox{#1}
  \parfillskip0pt\par
}
\newcommand\foo[1]{{\comfyfill{\mbox{#1}}}}
\renewcommand*{\backref}[1]{}
\renewcommand*{\backrefalt}[4]{%
\ifcase #1 %
\or
\foo{(page #2)}%
\else
\foo{(pages #2)}%
\fi
}

\usepackage{stringstrings}

\input{humble.tex}

\declareHumble{ANOVA}{ANOVA}
\declareHumble{ARD}{ARD}
\declareHumble{BIC}{BIC}
\declareHumble{BMC}{BMC}
\declareHumble{bq}{BQ}
\declareHumble{CRP}{CRP}
\declareHumble{dirpro}{DP}
\declareHumble{HDMR}{HDMR}
\declareHumble{GAM}{GAM}
\declareHumble{GEM}{GEM}
\declareHumble{GMM}{GMM}
\declareHumble{gplvm}{GP-LVM}
\declareHumble{gpml}{GPML}
\declareHumble{GPML}{GPML}
\declareHumble{gprn}{GPRN}
\declareHumble{gpt}{GP}
\declareHumble{gp}{GP}
\declareHumble{HKL}{HKL}
\declareHumble{HMC}{HMC}
\declareHumble{ibp}{IBP}
\declareHumble{iGMM}{iGMM}
\declareHumble{iwmm}{iWMM}
\declareHumble{kCP}{CP}
\declareHumble{kCW}{CW}
\declareHumble{kC}{C}
\declareHumble{KDE}{KDE}
\declareHumble{kLin}{Lin}
\declareHumble{KPCA}{KPCA}
\declareHumble{kPer}{Per}
\declareHumble{kPerGen}{ZMPer}
\declareHumble{kRQ}{RQ}
\declareHumble{kSE}{SE}
\declareHumble{kWN}{WN}
\declareHumble{Lin}{Lin}
\declareHumble{LBFGS}{L-BFGS}
\declareHumble{LIBSVM}{LIBSVM}
\declareHumble{MAP}{MAP}
\declareHumble{mcmc}{MCMC}
\declareHumble{MKL}{MKL}
\declareHumble{MLP}{MLP}
\declareHumble{MNIST}{MNIST}
\declareHumble{MSE}{MSE}
\declareHumble{OU}{OU}
\declareHumble{Per}{Per}
\declareHumble{RBF}{RBF}
\declareHumble{RMSE}{RMSE}
\declareHumble{RQ}{RQ}
\declareHumble{SBQ}{SBQ}
\declareHumble{seard}{SE-ARD}
\declareHumble{sefull}{SE-\textnormal{full}}
\declareHumble{SEGP}{SE-GP}
\declareHumble{SE}{SE}
\declareHumble{SNR}{SNR}
\declareHumble{SSANOVA}{SS-ANOVA}
\declareHumble{SVM}{SVM}
\declareHumble{UCI}{UCI}
\declareHumble{UMIST}{UMIST}
\declareHumble{vbgplvm}{VB GP-LVM}

\newcommand{\hphi}{h}
\newcommand{\hPhi}{\vh}

\newcommand{\lengthscale}{w}

\newcommand{\layerindex}{\ell}

\newcommand{\gpdrawbox}[1]{
\setlength\fboxsep{0pt}
\hspace{-0.15in} 
\fbox{
\includegraphics[width=0.464\columnwidth]{\deeplimitsfiguresdir/deep_draws/deep_gp_sample_layer_#1}
}}

\newcommand{\expectargs}[2]{\mathbb{E}_{#1} \left[ {#2} \right]}

\newcommand{\varianceargs}[2]{\mathbb{V}_{#1} \left[ {#2} \right]}
\newcommand{\cov}{\operatorname{cov}}

\newcommand{\colvec}[2]{\left[ \begin{array}{c} {#1} \\ {#2} \end{array} \right]}

\newcommand{\vect}[1]{{\bf{#1}}}
\newcommand{\mat}[1]{\mathbf{#1}}

\newcommand{\vb}{\vect{b}}

\newcommand{\vecf}{\boldsymbol{f}}

\newcommand{\vf}{\vecf}

\newcommand{\vh}{\vect{h}}

\newcommand{\vr}{\vect{r}}

\newcommand{\vV}{\mat{V}}
\newcommand{\vW}{\mat{W}}
\newcommand{\vw}{\vect{w}}

\newcommand{\vx}{\vect{x}}

\newcommand{\vy}{\vect{y}}
\newcommand{\vzero}{\vect{0}}

\newcommand{\netweights}{w}
\newcommand{\vnetweights}{\vw}
\newcommand{\mnetweights}{\vW}

\newcommand{\munitparams}{\vV}

\newcommand{\Nt}[2]{\mathcal{N}\!\left(#1,#2\right)}

\newcommand{\tra}{^{\mathsf{T}}}

\newcommand{\GPt}[2]{\mathcal{GP}\!\left(#1,#2\right)}

\newcolumntype{C}[1]{>{\centering\let\newline\\\arraybackslash\hspace{0pt}}m{#1}}
\newcolumntype{L}[1]{>{\raggedright\let\newline\\\arraybackslash\hspace{0pt}}m{#1}}
\newcolumntype{R}[1]{>{\raggedleft\let\newline\\\arraybackslash\hspace{0pt}}m{#1}}

\def\simiid{\overset{\mbox{\tiny iid}}{\sim}}
\def\simind{\overset{\mbox{\tiny \textnormal{ind}}}{\sim}}

\newcommand{\distas}[1]{\mathbin{\overset{#1}{\sim}}}

\def\GP{\mathcal{GP}}

\def\feat{\vh}

\theoremstyle{plain}
\newtheorem{theorem}{Theorem}[section]
\newtheorem{lemma}[theorem]{Lemma}

